\documentclass[11pt, logo, onecolumn, copyright, colorlinks=true, allcolors=blue]{nvidiatechreport}
\usepackage{graphicx} 
\usepackage{subcaption,ragged2e}
\usepackage[round]{natbib}

\usepackage[normalem]{ulem}
\usepackage[utf8]{inputenc} 
\usepackage[T1]{fontenc}    
\usepackage{url}
\usepackage{tikz}
\usepackage{enumitem}
\usetikzlibrary{positioning}
\usepackage{array}
\usepackage{booktabs}
\usepackage{wrapfig}
\usepackage{caption} 
\usepackage{listings}
\usepackage{adjustbox}
\usepackage{footnote}
\usepackage{multirow}
\usepackage{amsmath}
\usepackage{amsfonts}
\usepackage{breakcites}
\usepackage{blindtext}
\usepackage{svg}
\usepackage[most]{tcolorbox}
\usepackage{tabularx}
\usepackage{makecell}

\usepackage{tikz}
\usepackage{adjustbox} 

\usetikzlibrary{positioning,arrows.meta,backgrounds}

\definecolor{stagefill}{RGB}{226,241,214}   
\definecolor{bgblue}{RGB}{154,179,211}      
\definecolor{bgblue2}{RGB}{121,159,199}     
\usetikzlibrary{positioning,arrows.meta,backgrounds,fit}

\tikzset{
  >={Stealth[length=3mm,width=2mm]},
  stage/.style={
    draw, rounded corners=2pt, fill=stagefill,
    minimum width=4.2cm, minimum height=1.9cm,
    align=center, inner sep=6pt
  },
  band/.style={
    draw=none, fill=white, rounded corners=2pt,
    minimum height=8mm, align=center, inner sep=4pt
  },
  dashedgroup/.style={
    draw=black!50, dashed, rounded corners=3pt, inner sep=8pt
  }
}

\title{NVIDIA Nemotron Nano V2 VL}
\author{\large NVIDIA}
\date{}

\begin{document}

\begin{abstract}
\large \textbf{Abstract.}
We introduce Nemotron Nano V2 VL, the latest model of the Nemotron vision-language series designed for strong real-world document understanding, long video comprehension, and reasoning tasks. Nemotron Nano V2 VL delivers significant improvements over our previous model, Llama-3.1-Nemotron-Nano-VL-8B, across all vision and text domains through major enhancements in model architecture, datasets, and training recipes. Nemotron Nano V2 VL builds on Nemotron Nano V2, a hybrid Mamba-Transformer LLM, and innovative token reduction techniques to achieve higher inference throughput in long document and video scenarios. We are releasing model checkpoints in BF16, FP8, and FP4 formats and sharing large parts of our datasets, recipes and training code.
\end{abstract}

\maketitle

{\small
    {\href{https://huggingface.co/nvidia/NVIDIA-Nemotron-Nano-12B-v2-VL-BF16}{Model (BF16})} \quad|\quad
    {\href{https://huggingface.co/nvidia/NVIDIA-Nemotron-Nano-12B-v2-VL-FP8}{Model (FP8)}} \quad|\quad
    {\href{https://huggingface.co/nvidia/NVIDIA-Nemotron-Nano-12B-v2-VL-NVFP4-QAD}{Model (NVFP4-QAD)}} \quad|\quad
    {\href{https://huggingface.co/datasets/nvidia/Nemotron-VLM-Dataset-v2}{Dataset}}
}

\section{Introduction}
\label{section:introduction}

We introduce Nemotron Nano V2 VL, an efficient 12B vision–language model that achieves leading accuracy on OCRBench v2 \citep{fu2024ocrbenchv2improvedbenchmark} private data leaderboard\footnote{\url{https://99franklin.github.io/ocrbench_v2/}}, along with strong performance in reasoning, document understanding, long-video comprehension, visual question answering, and STEM reasoning. Nemotron Nano V2 VL delivers substantial improvements over our previous model, Llama-3.1-Nemotron-Nano-VL-8B\footnote{\url{https://huggingface.co/nvidia/Llama-3.1-Nemotron-Nano-VL-8B-V1}}, driven by enhancements in model architecture, dataset composition, and training methodology. These gains stem from the inclusion of higher-quality reasoning data, expanded OCR datasets, and additional long-context datasets.

In addition to overall benchmark improvements, we extend the model’s context length from 16K to 128K, enabling better handling of long videos and complex reasoning tasks. Consistent with our prior open-source efforts, we release the model weights, along with substantial portions of the training datasets, recipes, and codebase, to support continued research and development.

Nemotron Nano V2 VL builds on top of the Nemotron Nano V2 \citep{nvidia2025nvidianemotronnano2} 12B reasoning LLM and RADIOv2.5 vision encoder \citep{heinrich2025radiov25improvedbaselinesagglomerative}. In addition, Nemotron Nano V2 VL adopts the multimodal fusion architecture, training recipe as well as data strategy similar to Eagle 2 and 2.5 \citep{li2025eagle2buildingposttraining,chen2025eagle25boostinglongcontext}.  We use the open-source Megatron \citep{megatron-lm} framework to train the model in FP8 precision using Supervised Finetuning (SFT) across several vision and text domains, followed by additional SFT stages to further improve video understanding, long context performance, and recover text-only capabilities to achieve competitive results across many vision and text benchmarks. 

Compared to Llama-3.1-Nemotron-Nano-VL-8B, the hybrid Mamba-Transformer architecture of the LLM offers 35\% higher throughput in long multi-page document understanding scenarios. Additionally, we employ Efficient Video Sampling (EVS) \citep{bagrov2025efficientvideosamplingpruning} to accelerate throughput in video understanding use cases by 2x or more with minimal or no impact on accuracy. 

Furthermore, Nemotron Nano V2 VL supports both reasoning-on and reasoning-off modes, with the former enabling extended reasoning for tasks that require more complex problem-solving. This design enables a balanced trade-off between computational efficiency and task performance.

We are releasing our model weights on HuggingFace in BF16, FP8 and FP4 formats:
\begin{itemize}
    \item \href{https://huggingface.co/nvidia/NVIDIA-Nemotron-Nano-12B-v2-VL-BF16}{Nemotron-Nano-12B-v2-VL}: The final model weights after our multi-stage training recipe.
    \item \href{https://huggingface.co/nvidia/NVIDIA-Nemotron-Nano-12B-v2-VL-FP8}{Nemotron-Nano-12B-v2-VL-FP8}: Quantized model weights in FP8 format
    \item \href{https://huggingface.co/nvidia/NVIDIA-Nemotron-Nano-12B-v2-VL-NVFP4-QAD}{Nemotron-Nano-12B-v2-VL-NVFP4-QAD} Quantized model weights in FP4 format using Quantization-aware Distillation (QAD)
\end{itemize}

Additionally, we release a large portion of our SFT dataset and tooling:
\begin{itemize}
    \item \href{https://huggingface.co/datasets/nvidia/Nemotron-VLM-Dataset-v2}{Nemotron VLM Dataset V2}: A collection of over 8 million training samples.
    \item \href{https://github.com/NVIDIA-NeMo/Curator/tree/experimental/experimental/nvpdftex}{NVPDFTex}: A custom LaTeX compiler toolchain to generate annotated OCR ground truth.
\end{itemize}

The remainder of this report is organized as follows: Section \ref{section:model_arch} describes the model architecture and input processing pipeline; Section \ref{section:recipe} outlines the training recipe, datasets, and hyperparameters; and Section \ref{section:experiments} presents comprehensive evaluations of the model across both multimodal and pure-text tasks.
\section{Model Architecture}
\label{section:model_arch}

\begin{figure}[h!]
  \centering
  \includegraphics[width=\linewidth]{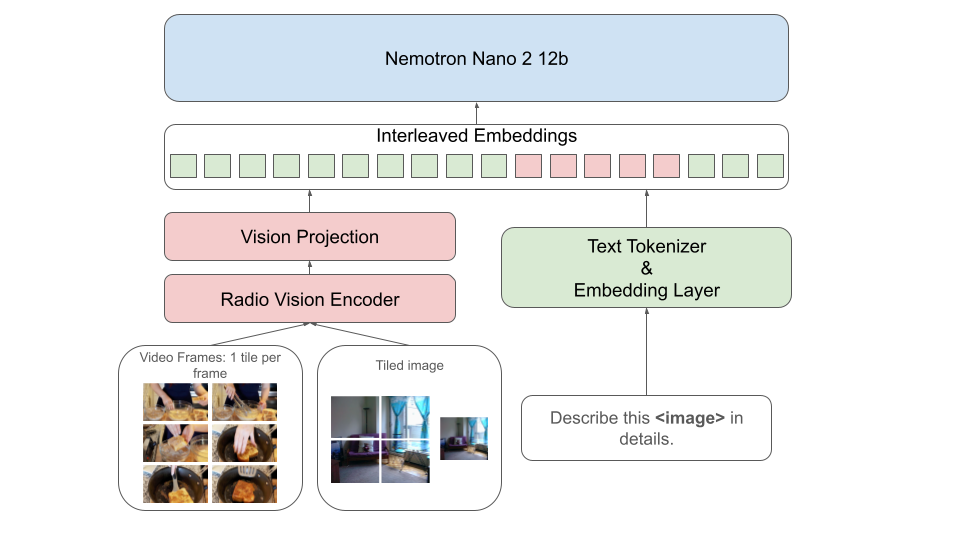} 
  \caption{Visualization of our VLM architecture. For images, we extract a dynamic number of tiles based on the image aspect ratio. For videos, we uniformly extract frames. Tiles and frames are resized to $512 \times 512$ pixels, and go through the RADIO vision encoder and an MLP connector. The image and text embeddings are interleaved, and fed to the Nemotron-Nano-12B-V2 LLM.}
  \label{fig:architecture}
\end{figure}

As illustrated in Figure ~\ref{fig:architecture}, Nemotron Nano V2 VL consists of three modules: a vision encoder, an MLP projector, and a language model. We initialize the vision encoder using the c-RADIOv2-VLM-H version of the RADIOv2 vision encoder \citep{heinrich2025radiov25improvedbaselinesagglomerative} and the language model with Nemotron-Nano-12B-V2 \citep{nvidia2025nvidianemotronnano2}.

Inspired by InternVL \citep{chen2024fargpt4vclosinggap}, LLaVA-1.5 \citep{liu2024improvedbaselinesvisualinstruction} and Eagle \citep{li2025eagle2buildingposttraining,chen2025eagle25boostinglongcontext}, we adopt a tiling strategy to handle varying image resolutions. First, each image is resized following the aspect ratio matching strategy employed by InternVL \citep{chen2024fargpt4vclosinggap} so that its width and height are multiples of $s$. Then it is divided into non-overlapping tiles of size $s\times s$. In this work, we set $s=512$. With a patch size of 16, this results in 1024 visual tokens per tile. For scalability, we employ pixel shuffle with 2x downsampling to reduce the token count further to 256. During training, we set the maximum number of tiles to 12. Additionally, we use a single-tile thumbnail of the image to capture global image context. For video inputs, we limit each input frame to a single tile.


\section{Training Recipe \& Datasets}
\label{section:recipe}

To preserve the base language model’s text comprehension and reasoning abilities while improving its visual understanding, we adopt a multi-stage training approach as detailed below.

\subsection{Stage 0}

In this stage, we aim to warm up the MLP connector to establish cross-modal alignment between the language and vision domains. To this end, we freeze the vision encoder and language model weights and train only the MLP connector on a diverse multimodal subset of the Stage 1 SFT dataset (see Section \ref{section:sft_stage_1}), consisting of approximately 2.2 million samples (up to 36 billion tokens) spanning multiple tasks, including captioning, visual question answering, visual grounding, OCR, and document extraction.

\subsection{SFT Stage 1: 16K context length}
\label{section:sft_stage_1}

In this and all subsequent stages, we unfreeze all model components for training. In SFT Stage 1, the maximum sequence length is set to 16,384 tokens. This stage is trained on approximately 32.5 million samples (about 112.5 billion tokens). To preserve the text comprehension capabilities of the Nemotron-Nano-V2 \citep{nvidia2025nvidianemotronnano2} LLM backbone, we incorporate a subset of the text reasoning data used in its Stage 1 SFT training, comprising approximately 6.5 million samples (around 40 billion tokens) spanning diverse domains and tasks such as mathematics, science, code, multilingual understanding, multi-turn dialogue, tool-use, and safety. In addition, we include multimodal datasets totaling 26 million samples (approximately 72 billion tokens) drawn from various tasks and sources, including:

\begin{enumerate}
    \item \textbf{Image Captioning}: OpenImages \citep{Kuznetsova_2020}, TextCaps \citep{sidorov2020textcapsdatasetimagecaptioning}, TextVQA \citep{singh2019towards}, PixMo-cap \citep{deitke2024molmopixmoopenweights}.
    \item \textbf{Video Captioning}: Localized Narratives \citep{PontTuset_eccv2020}, YouCook2 \citep{zhou2017automaticlearningproceduresweb}, VaTeX \citep{wang2020vatexlargescalehighqualitymultilingual}.
    \item \textbf{General Visual QA}: TextVQA \citep{singh2019towards}, VQAv2 \citep{goyal2017makingvvqamatter}, OK-VQA \citep{marino2019okvqavisualquestionanswering}, GQA \citep{hudson2019gqanewdatasetrealworld}, CLEVR \citep{johnson2016clevrdiagnosticdatasetcompositional}, CLEVR-Math \citep{lindström2022clevrmathdatasetcompositionallanguage}, TallyQA \citep{acharya2018tallyqaansweringcomplexcounting}, Dolly-15K \citep{DatabricksBlog2023DollyV2}, ScreenQA \citep{hsiao2025screenqalargescalequestionanswerpairs}, VizWiz \citep{gurari2018vizwizgrandchallengeanswering}, MapQA \citep{chang2022mapqadatasetquestionanswering}, ScienceQA \citep{lu2022learn}, PMC-VQA \citep{zhang2024pmcvqavisualinstructiontuning}, MetaMathQA \citep{yu2024metamathbootstrapmathematicalquestions}, UniGeo \citep{chen2022unigeounifyinggeometrylogical}, CMM-Math \citep{liu2024cmm}, Geo-170K \citep{gao2025gllavasolvinggeometricproblem}, VisualWebInstruct \citep{jia2025visualwebinstructscalingmultimodalinstruction}, LRV-Instruction \citep{liu2024mitigatinghallucinationlargemultimodal}, OCR-VQA \citep{8978122}, EST-VQA \citep{wang2020general}, ST-VQA \citep{biten2019scenetextvisualquestion}, PixMo-AskModelAnything \citep{deitke2024molmopixmoopenweights}, ALLaVA-4v \citep{chen2024allavaharnessinggpt4vsynthesizeddata}, SLAKE \citep{liu2021slakesemanticallylabeledknowledgeenhanceddataset}, VQA-RAD \citep{lau2018dataset}, DreamSim \citep{fu2023dreamsimlearningnewdimensions}, Spot-the-Diff \citep{jhamtani2018learningdifferencespairssimilar}, NLVR2 \citep{suhr2019corpusreasoningnaturallanguage}.
    \item \textbf{Video QA}: CLEVRER \citep{yi2020clevrercollisioneventsvideo}, Perception Test \citep{pătrăucean2023perceptiontestdiagnosticbenchmark}, ALFRED \citep{shridhar2020alfred}, NextQA \citep{xiao2021nextqanextphasequestionansweringexplaining}, VCG+112K \citep{Maaz2024VideoGPT+}.
    \item \textbf{Visual Grounding}: RefCOCO \citep{kazemzadeh-etal-2014-referitgame}.
    \item \textbf{OCR, Table \& Document Extraction}: SynthDog-en \citep{kim2022donut}, SynthTabNet \citep{nassar2022tableformertablestructureunderstanding}, DocLayNet \citep{Pfitzmann_2022}, WebSight \citep{laurençon2024unlockingconversionwebscreenshots}, TabRecSet \citep{yang2023large}, FinTabNet \citep{zheng2020globaltableextractorgte}, PubTables-1M \citep{smock2021pubtables1mcomprehensivetableextraction}, TextOCR \citep{singh2021textocr}, HierText \citep{long2022endtoendunifiedscenetext}, FUNSD \citep{jaume2019funsddatasetformunderstanding}, CASIA-HWDB2 \citep{liu2011casia}, RCTW-17 \citep{shi2018icdar2017competitionreadingchinese}, ReCTS-19 \citep{liu2019icdar2019robustreading}, human-annotated CommonCrawl \footnote{\label{footnote:commoncrawl}\url{https://commoncrawl.org/}} samples, synthetically generated tables, arXiv paper annotations generated using the NVPDFTex\footnote{\url{https://github.com/NVIDIA-NeMo/Curator/tree/experimental/experimental/nvpdftex}} pipeline and translated to several other languages using mBART-large-50 \citep{tang2020multilingual}, and multilingual Wikimedia\footnote{\url{https://dumps.wikimedia.org/}} dumps.
    \item \textbf{Document, Chart, Table and GUI QA}: ChartQA \citep{masry2022chartqabenchmarkquestionanswering}, InfoVQA \citep{mathew2021infographicvqa}, AI2D \citep{kembhavi2016diagramworthdozenimages}, DocVQA \citep{mathew2021docvqadatasetvqadocument}, FigureQA \citep{kahou2018figureqaannotatedfiguredataset}, ECD-10K \citep{yang2025effective}, ArXivQA \citep{li2024multimodalarxivdatasetimproving}, PlotQA \citep{methani2020plotqareasoningscientificplots}, PixMo-Docs \citep{deitke2024molmopixmoopenweights}, TabMWP \citep{lu2023dynamicpromptlearningpolicy}, SlideVQA \citep{tanaka2023slidevqadatasetdocumentvisual}, Docmatix \citep{laurençon2024buildingbetterunderstandingvisionlanguage}, DocReason25K \citep{hu2024mplugdocowl15unifiedstructure}, UniChart \citep{masry2023unichart}, SimChart9K \citep{xia2024structchartschemametricaugmentation}, MMTab \citep{zheng2024multimodaltableunderstanding}, VisText \citep{2023-vistext},  ScreenQA \citep{hsiao2025screenqalargescalequestionanswerpairs}, WaveUI-25K \citep{agentsea_waveui25k_2024}, as well as synthetic QA labels generated for FinTabNet \citep{zheng2020globaltableextractorgte}, HierText \citep{long2022endtoendunifiedscenetext} and CommonCrawl PDF samples transcribed using Nemo Retriever Parse\footnote{\url{https://build.nvidia.com/nvidia/nemoretriever-parse}} \citep{karmanov2025eclairextractingcontent}.
    \item \textbf{Visual Grounding}: Visual7W \citep{zhu2016visual7wgroundedquestionanswering}, OpenImages \citep{Kuznetsova_2020}.
    \item \textbf{Function Calling}: Glaive function calling \citep{glaive_function_calling_v2}, xLAM-60K \citep{liu2024apigenautomatedpipelinegenerating}.
\end{enumerate}

We augment the corpus with both human-annotated reasoning traces and model-generated traces produced by Qwen2.5-VL-32B-Instruct \citep{qwen2025qwen25technicalreport}, GLM-4.1V, and GLM-4.5V \citep{vteam2025glm45vglm41vthinkingversatilemultimodal} across multiple datasets to reinforce the extended reasoning ability of our model in reasoning-on mode. Additionally, for datasets lacking explicit QA labels, we generate synthetic question–answer pairs from existing OCR extractions or captions using LLMs from the Qwen2.5 \citep{qwen2025qwen25technicalreport} and Qwen3 \citep{yang2025qwen3technicalreport} families. A large portion of the training data in this stage is released at \href{https://huggingface.co/datasets/nvidia/Nemotron-VLM-Dataset-v2}{Nemotron VLM Dataset V2}.

\subsection{SFT Stage 2: 49K context extension}

In this stage, we extend the model’s context length to 49,152 tokens to enhance its capability for multi-image and video understanding. The model is trained on approximately 11M samples (around 55B tokens), including a subset of the Stage 1 dataset. We experimented with varying proportions of Stage 1 data and found that a 25\% reuse ratio offers a good balance between training efficiency and maintaining accuracy  across text, vision, multi-frame and video benchmarks.

In addition to the reused Stage 1 subset, we curate video and multi-image datasets comprising approximately 1.4 million samples (around 17 billion tokens), covering a diverse range of tasks across several data sources, including: (1) \textbf{Video Classification}: Kinetics \citep{carreira2018quovadisactionrecognition}; (2) \textbf{Dense Video Captioning}: YouCook2 \citep{zhou2017automaticlearningproceduresweb}, HiREST \citep{zala2023hierarchicalvideomomentretrievalstepcaptioning}, ActivityNet \citep{7298698}; (3) \textbf{Video Captioning}: EgoExoLearn \citep{huang2025egoexolearndatasetbridgingasynchronous}; (4) \textbf{Temporal Action Localization}: Breakfast Actions \citep{6909500}, Perception Test \citep{pătrăucean2023perceptiontestdiagnosticbenchmark}, HiREST \citep{zala2023hierarchicalvideomomentretrievalstepcaptioning}, HACS Segment \citep{zhao2019hacshumanactionclips}, FineAction \citep{liu2022fineactionfinegrainedvideodataset}, Ego4D-MQ \citep{grauman2022ego4dworld3000hours}, ActivityNet \citep{7298698}; (5) \textbf{Video Temporal Grounding}: YouCook2 \citep{zhou2017automaticlearningproceduresweb}, QuerYD \citep{oncescu2021querydvideodatasethighquality}, MedVidQA \citep{gupta2022datasetmedicalinstructionalvideo}, Ego4D-NLQ \citep{grauman2022ego4dworld3000hours}, DiDeMo \citep{hendricks2017localizingmomentsvideonatural}; (6) \textbf{General Video QA}: LLaVA-Video-178K \citep{zhang2025llavavideovideoinstructiontuning}, Ego4D \citep{grauman2022ego4dworld3000hours}, TVQA \citep{lei2019tvqalocalizedcompositionalvideo}, Perception Test \citep{pătrăucean2023perceptiontestdiagnosticbenchmark}, NextQA \citep{xiao2021nextqanextphasequestionansweringexplaining}, EgoExoLearn \citep{huang2025egoexolearndatasetbridgingasynchronous}, CLEVRER \citep{yi2020clevrercollisioneventsvideo}, and relabeling of the following datasets with Qwen2.5-VL-72B-Instruct \citep{bai2025qwen25vltechnicalreport} into MCQ and open-ended questions: TAPOS \citep{shao2020tapos}, HC-STVG \citep{tang2021humancentricspatiotemporalvideogrounding}, EgoProceL \citep{bansal2022viewbestviewprocedure}, CrossTask \citep{zhukov2019crosstaskweaklysupervisedlearning}; (7) \textbf{Multi-page QA}: Synthetic multi-page QA data constructed from CommonCrawl PDF documents using Nemo Retriever Parse extractions; and (8) \textbf{Multi-image captions}: Mementos \citep{wang2024mementoscomprehensivebenchmarkmultimodal}.

We convert all the non-QA data into QA formats. For video classification, temporal action localization and temporal grounding data, we use template questions to generate QA pairs. For video captioning and multi-page OCR captions, we use existing LLM models from the Qwen2.5 family \citep{qwen2025qwen25technicalreport} to synthesize both the questions and answers given the captions.

\subsection{SFT stage 3: 49K text recovery}

After SFT Stages 1 and 2, we observe a substantial drop in the LiveCodeBench score compared to the LLM backbone, despite including the text reasoning data from Nemotron Nano 2 \citep{nvidia2025nvidianemotronnano2}. To recover this loss, we introduce an additional SFT Stage 3 trained with a maximum sequence length of 49,152 using only code reasoning data totaling 1M samples or 15B tokens.

\subsection{SFT stage 4: 300K context extension}

Finally, we extend the model's context further and incorporate long-context data from the Stage 3 SFT stage of Nemotron Nano 2 \citep{nvidia2025nvidianemotronnano2}, accounting for around 74K samples or 12B tokens. The samples in this data are 160K tokens long on average, and we train with a maximum sequence length of 311,296 to accommodate the longest samples. We find that this stage helps improve the accuracy of the model in long-context benchmarks such as RULER \citep{hsieh2024rulerwhatsrealcontext}.

\subsection{Training details}

At all stages, the model is trained with FP8 precision following a recipe similar to \citep{nvidia2025nvidianemotronnano2} to accelerate training. The same configuration is applied to the LLM, vision encoder, and vision projection MLP, with the first and last layers of the LLM and the transformer blocks of the vision encoder kept in BF16. We did not observe any training instabilities with this setup, and both the training loss curve and benchmark scores closely match those of a full-BF16 model. Additionally, experiments keeping either the vision encoder or vision projection in BF16 did not yield a significant difference.

For video inputs to the model, we extract 2 frames per second, with a maximum of 128 frames for each video. If a video is longer than 64 seconds, we uniformly sample 128 frames instead.

As text-only, image and video samples vary significantly in sequence length, we find that sequence packing reduces training time by minimizing the number of padding tokens required for batching. We perform sequence packing online during training using a buffer containing several thousand samples. We employ the balance-aware data packing strategy described in \citep{li2025eagle2buildingposttraining}.

To mitigate any bias towards shorter or longer sequences during training, we employ loss square-averaging, similar to InternVL \citep{chen2025expandingperformanceboundariesopensource}. We use the AdamW optimizer with $\beta_1$ and $\beta_2$ set to $0.9$ and $0.999$, respectively, and a cosine annealing schedule with a linear warmup. See Table~\ref{tab:training_details} for an overview of the training hyperparameters.

\paragraph{Long Context Extension.} 

For SFT stages 2, 3, and 4, we employ context parallelism in the LLM \citep{MegatronContextParallelism}. Context parallelism partitions the LLM input along the sequence dimension, mitigating out-of-memory issues at longer sequence lengths. We use 2-way and 8-way context parallelism for SFT stages 2–3 and stage 4, respectively. When using N-way context parallelism in the LLM, N replicas of the vision encoder and vision projection modules are instantiated. To efficiently utilize these replicas and further reduce GPU memory usage, we follow the approach of \citep{chen2024longvilascalinglongcontextvisual} and split the vision encoder and projection inputs into N shards along the batch dimension. The N vision projection outputs are then gathered before being passed to the LLM.

\paragraph{Infrastructure.} We train the model with the Megatron \citep{megatron-lm} framework\footnote{\url{https://github.com/NVIDIA/Megatron-LM}} using Transformer Engine\footnote{\url{https://github.com/NVIDIA/TransformerEngine}} and the Megatron Energon\footnote{\url{https://github.com/NVIDIA/Megatron-Energon}} dataloader on NVIDIA H100 GPUs. Our training code is for large parts open-source. The training resources are summarized in Table~\ref{tab:training_resources}.

\begin{table}[t!]\small
\centering
\begin{tabular}{@{}l|ccccc@{}}
\toprule
& \textbf{Stage 0} & \textbf{Stage 1} & \textbf{Stage 2} & \textbf{Stage 3} & \textbf{Stage 4} \\ \toprule 
\texttt{Global Batch Size}  & 1024  & \multicolumn{4}{|c}{128} \\ \hline
\texttt{Total Training iterations} & 2158 & - & - & - & - \\ \hline
\texttt{Linear Warmup Fraction} & \multicolumn{5}{c}{0.1} \\ \hline
\texttt{Learning Rate} & $2\times 10^{-4}$ & \multicolumn{4}{|c}{$2\times 10^{-5}$} \\ \hline
\texttt{Weight Decay} & 0.01 & \multicolumn{4}{|c}{0.05} \\ \hline
\texttt{Max Length} & \multicolumn{2}{c|}{16384} & \multicolumn{2}{c|}{49152} & 311296 \\ 
\bottomrule
\end{tabular}
\caption{Training hyperparameters for all stages}
\label{tab:training_details}
\end{table}

\begin{table}[t!]\small
\centering
\begin{tabular}{@{}l|ccccc@{}}
\toprule
& \textbf{Stage 0} & \textbf{Stage 1} & \textbf{Stage 2} & \textbf{Stage 3} & \textbf{Stage 4} \\ \toprule
\texttt{\# GPU nodes}  & 32  & \multicolumn{4}{|c}{64} \\ \hline
\texttt{Training time (in hrs)} & 6 & 30 & 15 & 3.5 & 5.5 \\
\bottomrule
\end{tabular}
\caption{Hardware resources across training stages}
\label{tab:training_resources}
\end{table}

\begin{figure}[htbp]
  \centering
  \begin{adjustbox}{max width=\linewidth}
    \begin{tikzpicture}[font=\sffamily\small]
\node[stage] (pt) {\textbf{Stage 0: Pretraining}\\[1pt] 36B tokens\\ 16k context length};
\node[stage, right=9mm of pt] (sft) {\textbf{Stage 1: Img + Txt SFT}\\[1pt] 112.5B tokens\\ 16k context length};
\node[stage, right=9mm of sft] (vid) {\textbf{Stage 2: Video Context Extension}\\[1pt] 55B tokens\\ 49k context length};
\node[stage, right=9mm of vid] (lcb) {\textbf{Stage 3: Code Reasoning Healing}\\[1pt] 15B tokens\\ 49k context length};
\node[stage, right=9mm of lcb, minimum width=4.6cm] (lc) {\textbf{Stage 4: Long Context}\\[1pt] 12B tokens\\ 300k context length};

\draw[->, line width=0.9pt] (pt) -- (sft);
\draw[->, line width=0.9pt] (sft) -- (vid);
\draw[->, line width=0.9pt] (vid) -- (lcb);
\draw[->, line width=0.9pt] (lcb) -- (lc);

\begin{scope}[on background layer]
  \node[fill=bgblue, inner sep=6pt, fit=(pt)(lc)] {};
\end{scope}
\end{tikzpicture}
  \end{adjustbox}
  \caption{Overview of the training stages for the VLM along with the model context length and number of training tokens.}
  \label{fig:full-pipeline}
\end{figure}
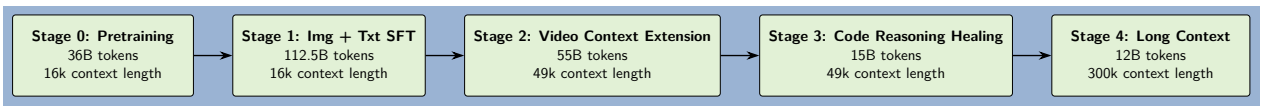
\section{Experiments}
\label{section:experiments}

To comprehensively assess our model’s capabilities, we evaluate it across a broad suite of multimodal benchmarks covering diverse tasks. In Section \ref{section:multimodal_evals}, we compare its performance with our previous-generation multimodal model, Llama-3.1-Nemotron-Nano-VL-8B, and with state-of-the-art multimodal models of similar scale. In Section \ref{section:text_only_evals}, we demonstrate the importance of our multi-stage training strategy in preserving the text reasoning abilities of the LLM backbone. We also investigate reasoning budget control and present the model’s behavior under different budget thresholds in Section \ref{section:budget_control}. In Section \ref{section:evs_video_eval}, we investigate the efficiency gains achieved by applying Efficient Video Sampling (EVS) \citep{bagrov2025efficientvideosamplingpruning} to video inputs. Finally, we examine alternative approaches for processing image inputs in Section \ref{section:tiling_vs_dynamic_res}.

\subsection{Multimodal Evaluations}
\label{section:multimodal_evals}

We conduct a comprehensive evaluation of our model on 45 benchmarks over seven broad categories:
\begin{enumerate}
    \item \textbf{General VQA}: MMBench V1.1 \citep{liu2024mmbenchmultimodalmodelallaround}, MMStar \citep{chen2024rightwayevaluatinglarge}, BLINK \citep{fu2024blink}, MUIRBench \citep{wang2024muirbenchcomprehensivebenchmarkrobust}, HallusionBench \citep{guan2024hallusionbenchadvanceddiagnosticsuite}, ZeroBench \citep{roberts2025zerobench}, CRPE \citep{wang2024allseeingprojectv2general}, POPE \citep{li2023evaluatingobjecthallucinationlarge}, MME-RealWorld \citep{zhang2025mmerealworldmultimodalllmchallenge}, RealWorldQA\footnote{\url{https://x.ai/news/grok-1.5v}}, MMT-Bench \citep{ying2024mmtbenchcomprehensivemultimodalbenchmark}, R-Bench \citep{guo2025rbenchgraduatelevelmultidisciplinarybenchmarks}, WildVision \citep{lu2024wildvision}. 
    \item \textbf{STEM Reasoning}: MMMU \citep{yue2023mmmu}, MMMU-Pro \citep{yue2025mmmuprorobustmultidisciplinemultimodal}, MathVista-Mini \citep{lu2024mathvista}, MathVision \citep{wang2024measuring}, MathVerse-Mini \citep{zhang2024mathversedoesmultimodalllm}, DynaMath \citep{zou2025dynamathdynamicvisualbenchmark}, LogicVista \citep{xiao2024logicvistamultimodalllmlogical}, WeMath \citep{qiao2024wemathdoeslargemultimodal}
    \item \textbf{Document Understanding, OCR \& Charts}: MMLongBench-Doc \citep{ma2024mmlongbenchdocbenchmarkinglongcontextdocument}, OCRBench \citep{Liu_2024}, OCRBench-V2 \citep{fu2024ocrbenchv2improvedbenchmark}, ChartQA \citep{masry2022chartqabenchmarkquestionanswering}, RDTableBench \citep{rdtablebench_dataset}, AI2D \citep{kembhavi2016diagramworthdozenimages}, TextVQA \citep{singh2019towards}, DocVQA \citep{mathew2021docvqadatasetvqadocument}, InfoVQA \citep{mathew2021infographicvqa}, OCR-Reasoning \citep{huang2025ocrreasoningbenchmarkunveilingtrue}, VCR \citep{zhang2025vcrtaskpixellevelcomplex}, SEED-Bench-2-Plus \citep{li2024seedbench2plusbenchmarkingmultimodallarge}, CharXiv \citep{wang2024charxivchartinggapsrealistic}
    \item \textbf{Visual Grounding \& Spatial Reasoning}: TreeBench \citep{wang2025traceableevidenceenhancedvisual}, CV-Bench \citep{tong2024cambrian1fullyopenvisioncentric}
    \item \textbf{GUI Understanding}: ScreenSpot \citep{cheng2024seeclickharnessingguigrounding}, ScreenSpot-v2 \citep{wu2024osatlasfoundationactionmodel}, ScreenSpot Pro \citep{li2025screenspotproguigroundingprofessional}
    \item \textbf{Video Understanding}: LongVideoBench \citep{wu2024longvideobenchbenchmarklongcontextinterleaved}, MLVU \citep{zhou2025mlvubenchmarkingmultitasklong}, Video-MME \citep{fu2025videommefirstevercomprehensiveevaluation}
    \item \textbf{Multimodal Multilingual Understanding}: MTVQA \citep{tang2025mtvqabenchmarkingmultilingualtextcentric}, MMMB \citep{sun2025parrotmultilingualvisualinstruction}, Multilingual MMBench \citep{sun2025parrotmultilingualvisualinstruction}
\end{enumerate}

\begin{table*}[htbp]
\centering
\resizebox{\textwidth}{!}{
\begin{tabular}{ll|cc|cc|cc|cc}
\toprule
\textbf{Task} & \textbf{Benchmark} & \multicolumn{2}{c|}{\textbf{Nemotron Nano V2 VL}} & \multicolumn{2}{c|}{\textbf{InternVL3.5}} & \multicolumn{2}{c|}{\textbf{GLM-4.5V}} & \multicolumn{2}{c}{\textbf{Qwen3-VL}} \\
\midrule 
Size & & 12B & 12B & 14B & 14B & 106B (A12B) & 106B (A12B) & 8B & 8B \\
Mode & & \makecell[c]{Reasoning\\off} & \makecell[c]{Reasoning\\on} & \makecell[c]{Non-\\Thinking} & Thinking & \makecell[c]{Non-\\Thinking} & Thinking & \makecell[c]{Instruct} & \makecell[c]{Thinking} \\
\midrule

\multirow{13}{*}{\textbf{General}} 
& MMBench V1.1 Dev (EN/ZH) & 83.0/80.2 & 82.6/76.3 & 83.0*/82.3* & - & - & - & 85.0/- & 87.5/- \\
& MMStar & 65.9 & 71.7 & 70.4 & - & 73.4 & 75.3 & 70.9 & 75.3 \\
& BLINK (Val) & 57.6 & 56.7 & 57.6 & - & 63.7 & 65.3 & 69.1 & 64.7  \\
& MUIRBench & 33.2 & 44.2 & 58.0 & - & 71.1 & 75.3 & 64.4 & 76.8 \\
& HallusionBench & 72.3 & 73.1 & 54.0 & - & 59.1 & 65.4 & 61.1 & 65.4\\
& ZeroBench (sub) & 15.0 & 18.3 & 10.8* & - & 21.9 & 23.4 & 22.8 & 18.6* \\
& CRPE (Relation) & 71.3 & 64.3 & 76.6 & - & - & - & 71.7* & 52.3* \\
& POPE & 88.8 & 87.0 & 87.7 & - & - & - & 88.2* & 86.2* \\
& MME-RealWorld (EN) & 62.2 & - & 63.2 & - & - & - & - & - \\
& RealWorldQA & 76.2 & 72.2 & 70.5 & - & - & - & 71.5 & 73.5 \\
& MMT-Bench (Val) & 65.5 & 63.8 & 66.9* & - & - & - & - & - \\
& R-Bench (dis) & 75.2 & 69.7 & 70.9 & - & - & - & 70.1* & 69.3* \\
& WildVision (win rate) & 16.0 & 20.2 & 73.0 & - & - & - & 75.0* & 76.4*  \\
\midrule

\multirow{8}{*}{\makecell[l]{\textbf{STEM}\\\textbf{Reasoning}}} 
& MMMU (val) & 55.3 & 67.8 & 62.78* & 73.3 & 68.4 & 75.4 & 69.6 & 74.1 \\
& MMMU Pro & 14.5 & 28.0 & - & - & 59.8 & 65.2 & 55.6 & 60.4 \\
& MathVista-Mini & 69.0 & 75.5 & 73.3* & 80.5 & 78.2 & 84.6 & 77.2 & 81.4  \\
& MathVision & 31.5 & 53.6 & - & 59.9 & 52.5 & 65.6 & 53.9 & 62.7 \\
& MathVerse-Mini (Vision-only) & 34.3 & 58.2 & 41.75* & 62.8 & - & - & 38.2* & 68.7* \\
& Dynamath (worst case) & 14.8 & 32.3 & - & 38.7 & 44.1 & 53.9 & 38.7* & - \\
& LogicVista & 38.7 & 58.4 & - & 60.2 & 54.8 & 62.4 & 58.8* & 59.3* \\
& WeMath & 31.5 & 39.3 & - & 58.7 & 58.9 & 68.8 & 57.1* & - \\
\midrule

\multirow{13}{*}{\makecell[l]{\textbf{Document}\\\textbf{Understanding,}\\\textbf{OCR \& Charts}}} 
& MMLongBench-Doc & 32.1 & - & - & - & 41.1 & 44.7 & 47.9 & 48.0 \\
& OCRBench & 85.6 & 83.5 & 83.6 & - & 87.2 & 86.5 & 89.6 & 81.9 \\
& OCRBenchV2 (EN/ZH) & 62.0/44.2 & 54.8/39.8 & - & - & - & - & 65.4/61.2 & 63.9/59.2 \\
& ChartQA (Test) & 89.8 & 84.9 & 86.5 & - & - & - & 83.4* & - \\
& RDTableBench & 67.8 & 57.4 & - & - & - & - & 82.4* & 43.9* \\
& AI2D (Test) & 87.2 & 84.7 & 85.1  & - & - & - & 85.7 & 84.9 \\
& TextVQA (Val) & 85.4 & 76.1 & 77.8 & - & - & - & 82.9* & - \\
& DocVQA (Test) & 94.7 & 93.2 & 93.4 & - & - & - & 96.1 & 95.3 \\
& InfoVQA (Test) & 79.4 & 80.4 & 78.3 & - & - & - & 83.1 & 86.0 \\
& OCR-Reasoning & 21.0 & 33.9 & - & - & - & - & 39.3* & 48.4* \\
& VCR-EN-Easy (EM/Jaccard) & 71.8/80.0 & - & 93.4/97.7 & - & - & - & 91.8*/95.1* & 87.4*/94.4*  \\
& SEED-Bench-2-Plus & 71.3 & 70.0 &  70.7 & - & - & - & 71.4* & 69.9* \\
& CharXiv (RQ/DQ) & 41.7/76.5 & 41.3/77.2 &  47.9/76.6 & - & - & - & 46.4/83.0 & 53.0/85.9 \\
\midrule

\multirow{2}{*}{\makecell[l]{\textbf{Visual Grounding \&}\\\textbf{Spatial Reasoning}}}
& TreeBench & 38.5 & 42.5 & 42.0* & - & 47.9 & 50.1 & 26.9* & 25.4* \\
& CV-Bench & 81.0 & 78.3 & 80.1* & - & 86.5 & 87.3 & 73.1* & 79.5* \\
\midrule

\multirow{3}{*}{\textbf{GUI}} 
& ScreenSpot & 39.4 & 40.1 & 87.5 & - & - & - & 94.4 & 93.6 \\
& ScreenSpot-v2 & 41.7 & 42.8 & 88.6 & - & - & - & - & -  \\
& ScreenSpot-Pro & 4.8 & 5.5 & - & - & - & - & 54.6 & 46.6 \\
\midrule

\multirow{3}{*}{\makecell[l]{\textbf{Video}\\\textbf{Understanding}}} 
& LongVideoBench & 63.6 & 57.0 & 62.7 & - & - & - & - & - \\
& MLVU (M-Avg) & 73.6 & - & 72.1 & - & - & - & - & - \\
& VideoMME (w/o sub) & 66.0 & 63.0 & 67.9 & - & 74.3 & 74.6 & 71.4 & 71.8 \\

\bottomrule
\end{tabular}
}
\caption{Comparison of Nemotron Nano V2 VL with existing open-source multimodal models. All results marked with * were calculated in VLMEvalKit.}
\label{tab:vlm-benchmark-summary}
\end{table*}

We use the VLMEvalKit \citep{duan2025vlmevalkitopensourcetoolkitevaluating}\footnote{\url{https://github.com/open-compass/VLMEvalKit}} framework for our evaluations with a vLLM \citep{kwon2023efficient}\footnote{\url{https://github.com/vllm-project/vllm}} backend inference server. For the reasoning-off mode, we employ greedy decoding and cap the maximum number of generated tokens at 1,024 for all benchmarks except RDTableBench\footnote{\url{https://huggingface.co/datasets/reducto/rd-tablebench/tree/main}} where we use a limit of 16,384 tokens. For reasoning-on evaluations, we set the temperature to 0.6, top-p to 0.95, and the maximum output length to 16,384 tokens.

We present reasoning-off evaluation results on a subset of the benchmarks after each training stage in Table \ref{tab:vision-across-stages}, along with a comparison to our previous generation multimodal model, Llama-3.1-Nemotron-Nano-VL-8B. We also compare our model against state-of-the-art open-source multimodal models of similar scale including InternVL3.5 (14B) \citep{wang2025internvl35advancingopensourcemultimodal}, GLM-4.5V (106B-A12B) \citep{vteam2025glm45vglm41vthinkingversatilemultimodal} and Qwen3-VL (8B) \citep{yang2025qwen3technicalreport}\footnote{\url{https://github.com/QwenLM/Qwen3-VL}} in Tables \ref{tab:vlm-benchmark-summary} and \ref{tab:vlm-multilingual}. Unless otherwise noted, we report the evaluation scores directly from the respective model reports. For benchmarks not covered therein, we independently evaluate the models using VLMEvalKit whenever possible.

\begin{table*}[htbp]
\centering
\scriptsize
\setlength{\tabcolsep}{4pt}
\renewcommand{\arraystretch}{1.1}
\begin{adjustbox}{width=\textwidth}
\begin{tabular}{l |*{5}{c}|c}
\toprule
\multicolumn{1}{l}{\textbf{Benchmark}} &
\multicolumn{1}{|c}{\textbf{\shortstack[c]{Stage 0}}} &
\multicolumn{1}{c}{\textbf{\shortstack[c]{SFT Stage 1:\\16K context\\length}}} &
\multicolumn{1}{c}{\textbf{\shortstack[c]{SFT Stage 2:\\49K context\\extension}}} &
\multicolumn{1}{c}{\textbf{\shortstack[c]{SFT stage 3:\\49K text\\recovery}}} &
\multicolumn{1}{c}{\textbf{\shortstack[c]{SFT stage 4:\\300K context\\extension}}} &
\multicolumn{1}{|c}{\textbf{\shortstack[c]{Llama-3.1-Nemotron\\Nano-VL-8B}}} \\
\midrule
\texttt{AI2D (Test)}                 & 67.6 & 87.1 & 87.1 & \textbf{87.3} & 87.2 & 85.0 \\
\texttt{ChartQA (Test)}              & 70.9 & 89.9 & 90.0 & \textbf{90.2}  & 89.8 & 86.3 \\
\texttt{DocVQA (Val)}                & 79.1 & \textbf{94.4} & 94.3 & 94.2 & \textbf{94.4} & 91.2 \\
\texttt{InfoVQA (Val)}               & 51.3 & \textbf{80.2} & 79.2 & 79.2 & 79.2 & 77.4 \\
\texttt{LongVideoBench}             & 45.1 & 59.4 & \textbf{63.6} & 63.1 & \textbf{63.6} & - \\
\texttt{MMLongBench-DOC}     & 10.75 & 29.2 & 32.0 & 30.8 & \textbf{32.1} & - \\
\texttt{MMMU (Val)}             & 49.0 & 54.8 & 55.0 & 54.3 & \textbf{55.3} & 48.2 \\
\texttt{MathVista-Mini}            & 53.1 & 67.7 & \textbf{69.3} & 69.2 & 69.0 & - \\
\texttt{OCRBench}                   & 61.4 & 84.8 & 85.3 & 85.4 & \textbf{85.6} & 83.9 \\
\texttt{OCRBench-V2 (CN)}              & 18.3 & 42.9 & 43.4 & 43.4 & \textbf{44.2} & 37.9 \\
\texttt{OCRBench-V2 (EN)}              & 38.3 & \textbf{62.5} & \textbf{62.5} & 61.8 & 62.0 & 60.1 \\
\texttt{TextVQA (Val)}               & 76.7 & \textbf{86.0} & 85.0 & 85.2 & 85.4 & - \\
\texttt{Video-MME}                  & 36.3 & 57.6 & 65.8 & 65.4 & \textbf{66.0} & 54.7 \\
\bottomrule
\end{tabular}
\end{adjustbox}
\caption{Vision benchmarks for our previous-generation VLM, Llama-3.1-Nemotron-Nano-VL-8B, and after each training stage of Nemotron Nano V2 VL with reasoning-off.}
\label{tab:vision-across-stages}
\end{table*}

\paragraph{Evaluation across training stages.} Table \ref{tab:vision-across-stages} highlights the effectiveness of our long-context extension training introduced in SFT stages 2 and 4. We observe substantial gains on Video-MME \citep{fu2025videommefirstevercomprehensiveevaluation} and MMLongBench-Doc \citep{ma2024mmlongbenchdocbenchmarkinglongcontextdocument} following SFT stage 2, and these improvements are retained in the final model after stage 4 training.

\paragraph{Comparison with Llama-3.1-Nemotron-Nano-VL-8B.} As shown in Table \ref{tab:vision-across-stages}, we observe consistent improvements across all benchmarks compared to Llama-3.1-Nemotron-Nano-VL-8B. We attribute these gains to a combination of factors, including an enhanced LLM backbone, expanded and higher-quality training datasets, and an improved training recipe.

\begin{table*}[htbp]
\centering
\resizebox{\textwidth}{!}{
\begin{tabular}{ll|c|c|c|c}
\toprule
\textbf{Task} & \textbf{Benchmark} & \makecell[c]{\textbf{Nemotron Nano V2 VL}\\\textbf{Reasoning-off}} & \makecell[c]{\textbf{InternVL3.5}\\\textbf{Non-Thinking}} & \textbf{GLM-4.5V} & \makecell[c]{\textbf{Qwen3-VL}\\\textbf{Instruct}} \\
\midrule 
Size & & 12B & 14B & 106B (A12B) & 8B \\
\midrule

\textbf{MTVQA} & Avg & 24.3 & 34.2 & 22.0$^{\dagger}$ & 32.2* \\
\midrule

\multirow{6}{*}{\textbf{MMMB}} 
& en & 86.1 & 85.1 & 87.1$^{\dagger}$ & 85.4* \\
& zh & 84.1 & 84.1 & 86.9$^{\dagger}$ & 82.5* \\
& pt & 83.5 & 82.7 & 84.8$^{\dagger}$ & 81.5* \\
& ar & 83.4 & 80.3 & 84.5$^{\dagger}$ & 80.3* \\
& tr & 77.4 & 79.4 & 84.6$^{\dagger}$ & 78.7* \\
& ru & 83.9 & 83.5 & 84.3$^{\dagger}$ & 81.9* \\
\midrule

\multirow{6}{*}{\makecell[l]{\textbf{Multilingual}\\\textbf{MMBench}}} 
& en & 84.9 & 84.0 & 89.1$^{\dagger}$ & 85.5* \\
& zh & 82.5 & 83.7 & 89.3$^{\dagger}$ & 85.6* \\
& pt & 82.5 & 80.0 & 86.9$^{\dagger}$ & 82.5* \\
& ar & 81.5 & 77.8 & 83.7$^{\dagger}$ & 79.0* \\
& tr & 75.7 & 77.0 & 84.0$^{\dagger}$ & 79.0* \\
& ru & 82.2 & 77.0 & 87.2$^{\dagger}$ & 81.3* \\

\bottomrule
\end{tabular}
}
\caption{Comparison of Nemotron Nano V2 VL with SOTA multimodal models on multimodal multilingual benchmarks. All scores marked by * were reproduced by us using VLMEvalKit, and those marked with $^{\dagger}$ were obtained from the InternVL3.5 \citep{wang2025internvl35advancingopensourcemultimodal} technical report.}
\label{tab:vlm-multilingual}
\end{table*}


\subsection{Pure Text Evaluations}
\label{section:text_only_evals}

We conduct all pure-text evaluations using the NeMo-Skills\footnote{\url{https://github.com/NVIDIA-NeMo/Skills}} framework with a maximum output length of 32,768 tokens, temperature set to 0.6, and top-p of 0.95. We report Pass@1 average of 16 runs for AIME-2025; an average of 4 runs for MATH-500 \citep{lightman2023letsverifystepstep}, GPQA-Diamond \citep{rein2023gpqagraduatelevelgoogleproofqa}, LiveCodeBench (07/24 - 12/24) \citep{jain2024livecodebenchholisticcontaminationfree}, IFEval \citep{zhou2023instructionfollowingevaluationlargelanguage}; and score of 1 run for SciCode \citep{tian2024scicoderesearchcodingbenchmark} and RULER \citep{hsieh2024rulerwhatsrealcontext}. For MMLU-Pro \citep{wang2024mmluprorobustchallengingmultitask}, we report the accuracy on a 1000-sample subset.

\begin{table*}[htbp]
\centering
\scriptsize
\setlength{\tabcolsep}{4pt}
\renewcommand{\arraystretch}{1.1}
\begin{adjustbox}{width=\textwidth}
\begin{tabular}{l *{5}{c}}
\toprule
\multicolumn{1}{c}{\textbf{Benchmark}} &
\multicolumn{1}{c}{\textbf{\shortstack[c]{Stage 0:\\Pretraining}}} &
\multicolumn{1}{c}{\textbf{\shortstack[c]{SFT Stage 1:\\16K context length}}} &
\multicolumn{1}{c}{\textbf{\shortstack[c]{SFT Stage 2:\\49K context extension}}} &
\multicolumn{1}{c}{\textbf{\shortstack[c]{SFT stage 3:\\49K text recovery}}} &
\multicolumn{1}{c}{\textbf{\shortstack[c]{SFT stage 4:\\300K context extension}}} \\
\midrule
\texttt{MATH-500}                    & 97.7 & 96.8 & 97.3 & 97.6 & 96.9 \\
\texttt{AIME-25}                      & 75.9 & 68.0 & 72.7 & 72.7 & 71.3 \\
\texttt{GPQA}                        & 65.0 & 60.9 & 63.0 & 60.6 & 64.1 \\
\texttt{LiveCodeBench}               & 70.0 & 50.9 & 55.0 & 69.8 & 69.4 \\
\texttt{IFEval\_prompt\_strict}      & 84.2 & 77.5 & 77.3 & 76.5 & 78.2 \\
\texttt{IFEval\_instruction\_strict} & 89.3 & 84.1 & 83.9 & 83.4 & 84.7 \\
\texttt{SciCode\_problem\_accuracy}  & 7.5  & 5.0  & 6.9  & 5.0  & 6.9  \\
\texttt{SciCode\_subtask\_accuracy}  & 22.3 & 17.5 & 14.9 & 17.2 & 17.6 \\
\texttt{MMLU-Pro-1000}               & 77.8 & 75.2 & 75.8 & 76.7 & 77.1 \\
\texttt{RULER}                       & 77.9 & 8.8  & 17.4 & 21.5 & 72.1 \\
\bottomrule
\end{tabular}
\end{adjustbox}
\caption{Text benchmarks with reasoning on for the different stages. Our goal was to add vision capabilities  with minimal impact to the text reasoning capabilities of the underlying LLM Nemotron-Nano-12B-V2. The text reasoning benchmarks of Stage 0 corresponds to the benchmarks results of the underlying LLM since the LLM is frozen during Stage 0. After Stage 1, we see a significant drop in text reasoning benhchmarks (eg. LiveCodeBench goes from 70 to 50.87) and long context benchmark (RULER goes from 77.91 to 8.80). The video long context extension (Stage 2) helps recover a little bit of the benchmark score, but we still observe very low LiveCodeBench (55.00) and RULER (17.39). We apply a code reasoning recovery stage (Stage 3), and a long context extension stage (Stage 4), to recover both benchmarks (LiveCodeBench of 69.44 and RULER of 72.12).}
\label{tab:text-across-stages}
\end{table*}

\paragraph{Evaluation across training stages.} We report the model’s text evaluation scores after each training stage in Table \ref{tab:text-across-stages}. Comparing the results between stage 0 (where the LLM backbone remains frozen) and SFT stage 1, we observe a significant drop in the LiveCodeBench score - from 70.0 to 50.87. To address this degradation without compromising vision performance, we explored several mitigation strategies that were unsuccessful, including augmenting the SFT stage 1 dataset with additional code reasoning examples and disabling loss scaling. Ultimately, we introduced an additional SFT stage 3 focused exclusively on code reasoning, which helped restore the LiveCodeBench score without affecting other benchmarks. As shown in Table \ref{tab:vision-across-stages}, the vision benchmarks remain stable between SFT stage 2 and stage 3, and the final model largely preserves the text reasoning capabilities of the original LLM backbone across most tasks.

Furthermore, we see a significant improvement in the RULER score across SFT stages 2-4 following an initial drop after stage 1. These results further validate the effectiveness of our long-context training strategy.

\subsection{Reasoning Budget Control}
\label{section:budget_control}

\begin{figure}[ht!]
  \centering
  \begin{subfigure}[t]{0.49\textwidth}
    \centering
    \includegraphics[width=\linewidth]{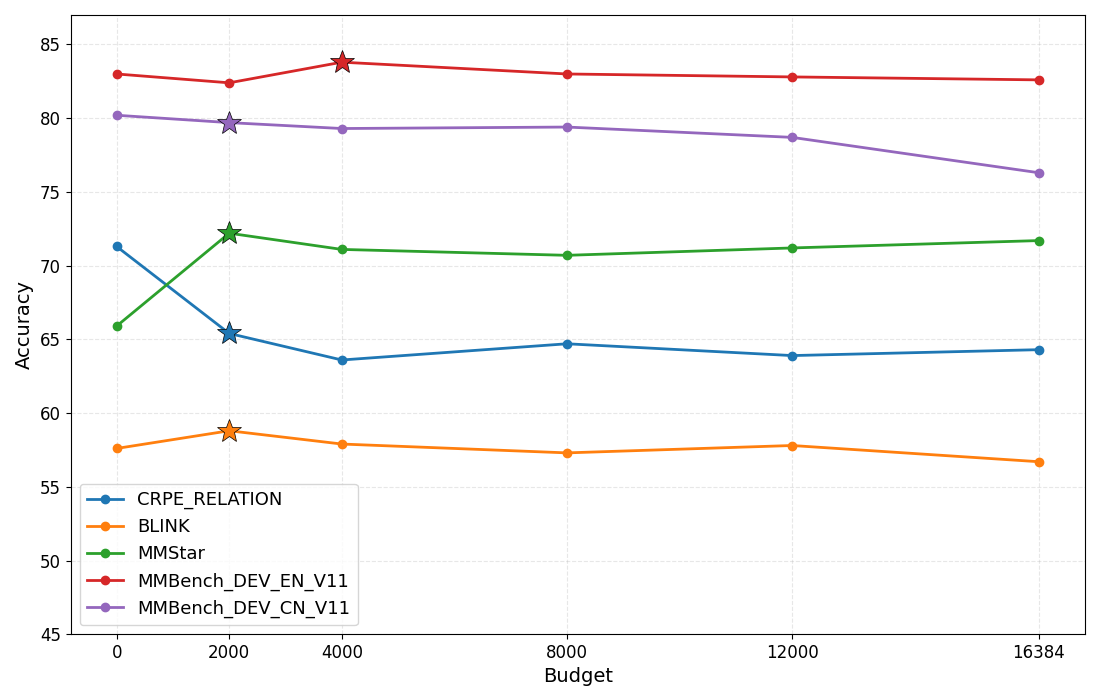}
    \subcaption{General VQA}\label{fig:five:a}
  \end{subfigure}
  \begin{subfigure}[t]{0.49\textwidth}
    \centering
    \includegraphics[width=\linewidth]{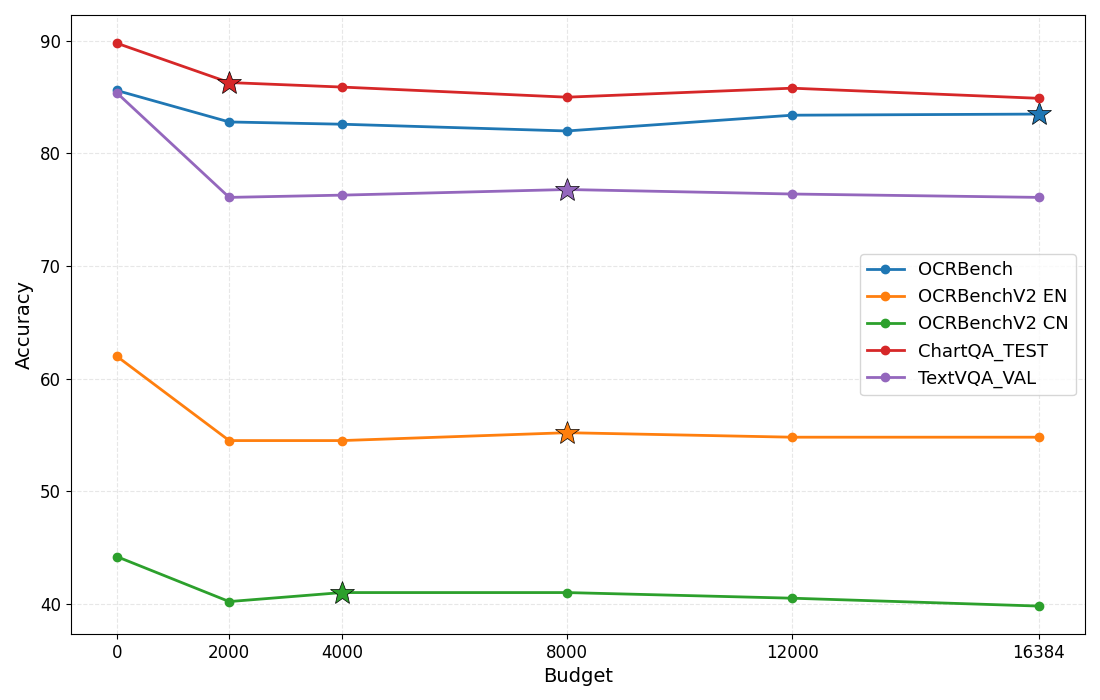}
    \subcaption{Document, OCR \& Chart Understanding}\label{fig:five:b}
  \end{subfigure}

  \vspace{0.8em}

  \begin{subfigure}[t]{0.49\textwidth}
    \centering
    \includegraphics[width=\linewidth]{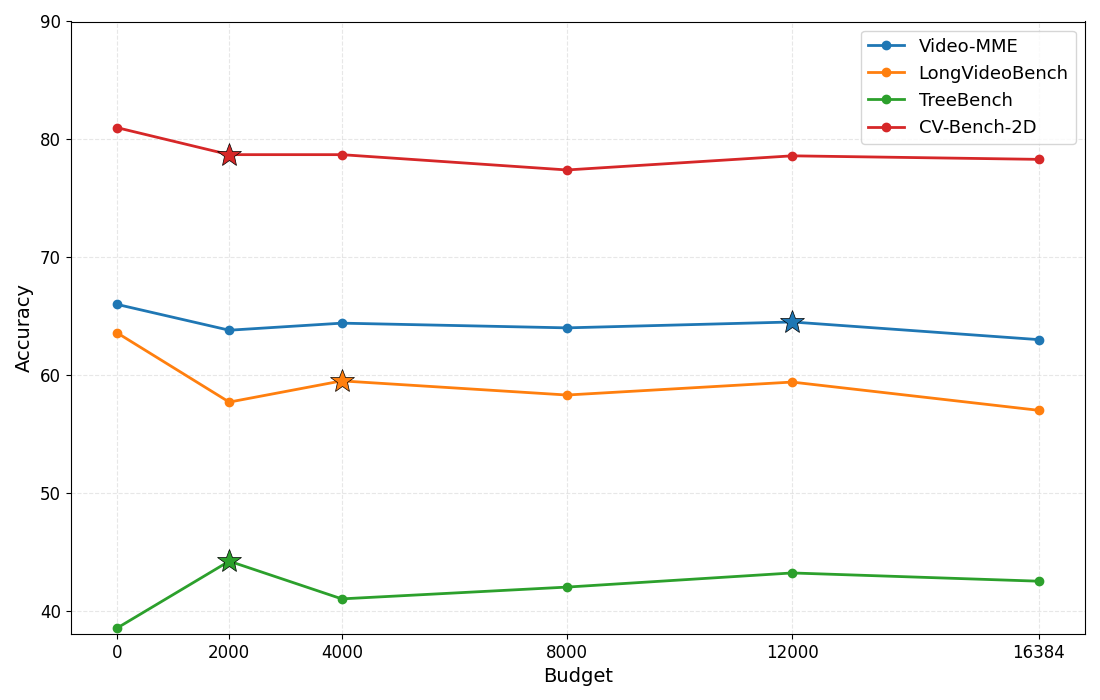}
    \subcaption{Spatial \& Video Understanding}\label{fig:five:d}
  \end{subfigure}
  \begin{subfigure}[t]{0.49\textwidth}
    \centering
    \includegraphics[width=\linewidth]{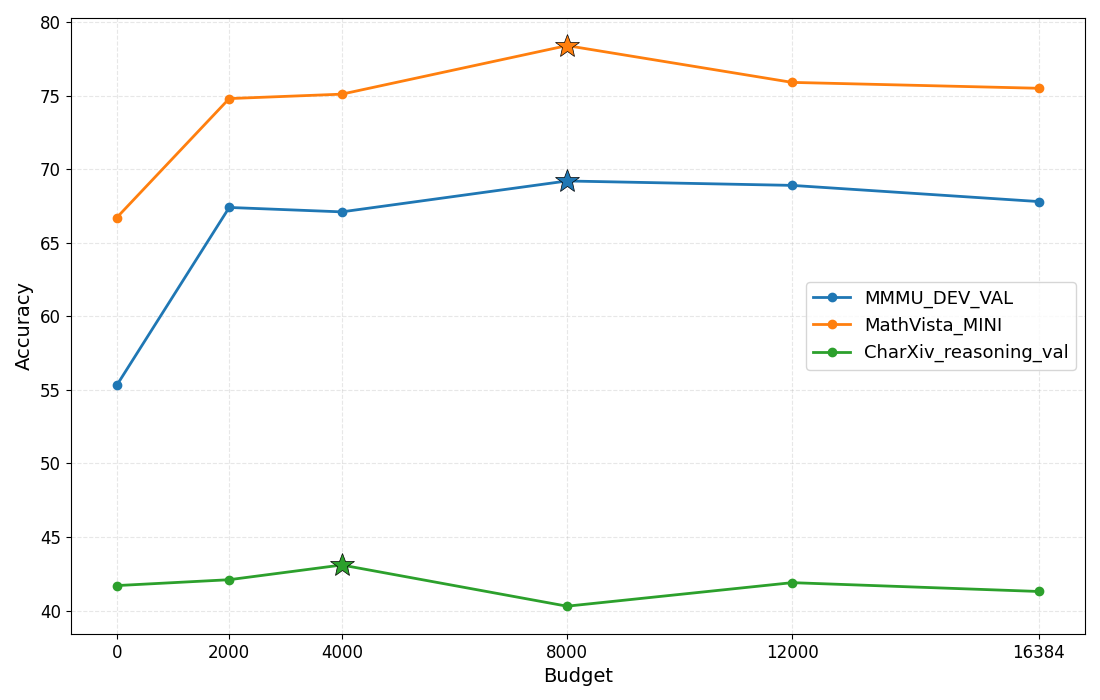}
    \subcaption{STEM \& Chart Reasoning}\label{fig:five:e}
  \end{subfigure}

  \caption{Effect of reasoning budget control at 2K, 4K, 8K, and 12K tokens across multiple tasks. A budget value of 0 corresponds to reasoning-off evaluations, while 16,384 denotes unrestricted reasoning-on evaluations with a maximum generation length of 16,384 tokens. The highest reasoning-on score (including the unrestricted case) is indicated with a $\bigstar$ for each task.}
  \label{fig:budget-control}
\end{figure}

Following Nemotron-Nano-V2 \citep{nvidia2025nvidianemotronnano2}, we experiment with varying reasoning budgets during inference. We evaluate the model’s behavior under budgets of 2K, 4K, 8K, and 12K tokens, each with a 500-token grace period, as shown in Figure \ref{fig:budget-control}.

Our experiments indicate that tuning the reasoning budget can improve reasoning-on mode accuracy across multiple tasks, even in cases where the unrestricted reasoning-on score is lower than the reasoning-off score. These gains with budget control may arise from the early termination of malformed reasoning traces with repetition loops on out-of-distribution tasks, as well as the truncation of overly verbose reasoning chains for problems requiring minimal or straightforward reasoning.

\subsection{Inference with Efficient Video Sampling}
\label{section:evs_video_eval}

Building upon \textbf{Efficient Video Sampling (EVS)} \citep{bagrov2025efficientvideosamplingpruning}, we integrate it directly into our video-processing pipeline.  
EVS reduces the number of visual tokens by identifying and pruning \emph{temporally static patches} -- spatial regions that remain nearly unchanged between consecutive frames, while preserving positional identity and semantic consistency.  
This enables our model to process substantially longer videos with lower latency and memory consumption, without requiring architectural changes or retraining.

\begin{figure*}[htbp]
  \centering
  \begin{minipage}[t]{0.48\textwidth}
    \vspace{20pt}
    \centering
    \begin{adjustbox}{width=\textwidth}
    \begin{tabular}{lcccc}
    \toprule
    \textbf{EVS} & \textbf{LongVideoBench} & \textbf{Video-MME} & \textbf{TTFT} & \textbf{Throughput} \\
    & & & \textbf{(ms)} & \textbf{(tok/s)} \\
    \midrule
    OFF  & 63.6 & 66.0 & 4131 & 34 \\
    50\% & 63.7 & 66.0 & 2699 & 65 \\
    60\% & 63.6 & 66.0 & 2458 & 75 \\
    70\% & 62.2 & 65.7 & 2184 & 84 \\
    75\% & 62.5 & 66.1 & 2072 & 88 \\
    80\% & 62.4 & 65.6 & 1990 & 98 \\
    90\% & 60.7 & 64.0 & 1654 & 120 \\
    \bottomrule
    \end{tabular}
    \end{adjustbox}
  \end{minipage}%
  \hfill
  \begin{minipage}[t]{0.48\textwidth}
    \vspace{0pt}
    \centering
    \includegraphics[width=\linewidth]{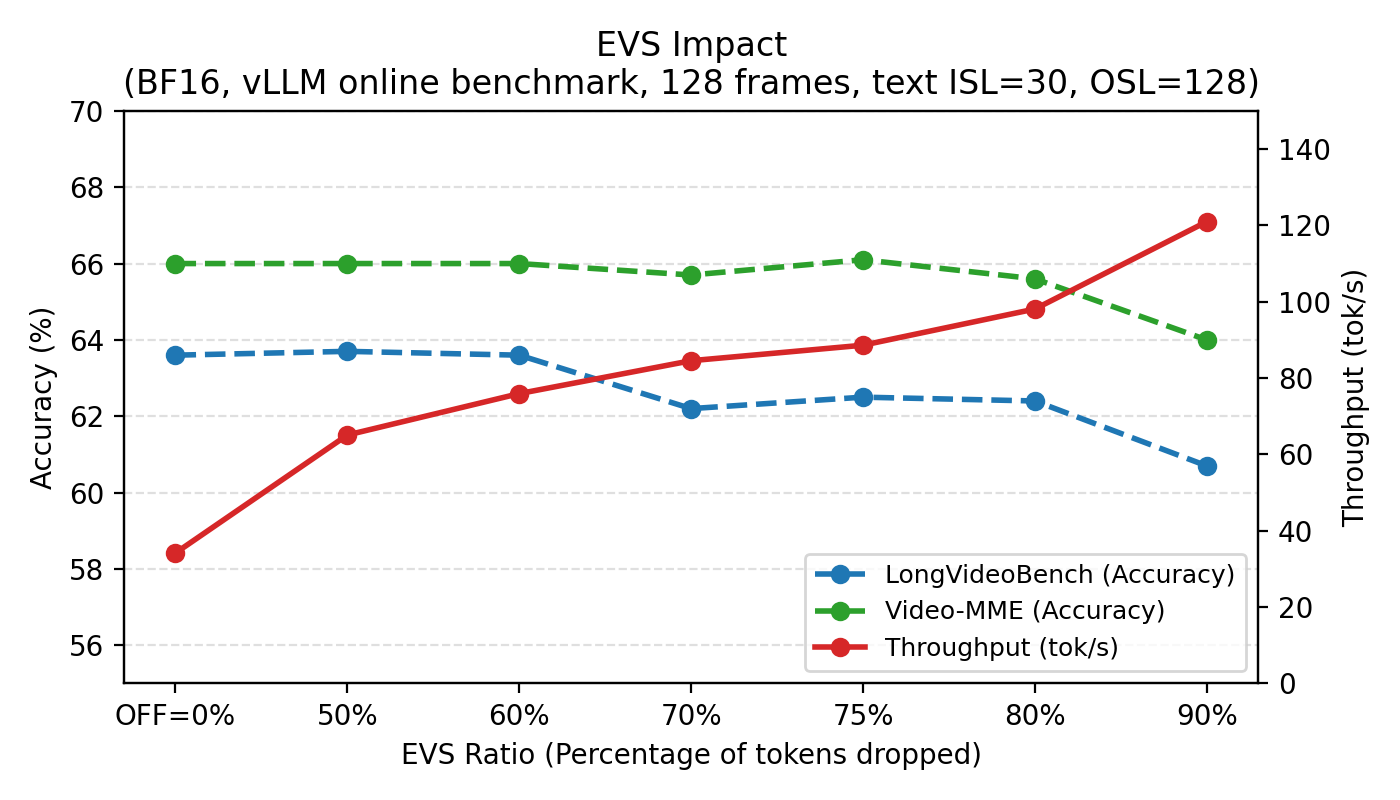}
  \end{minipage}

  \vspace{1em} 

  \begin{minipage}[t]{0.48\textwidth}
    \vspace{20pt}
    \centering
    \begin{adjustbox}{width=\textwidth}
    \begin{tabular}{lcccc}
    \toprule
    \textbf{EVS} & \textbf{LongVideoBench} & \textbf{Video-MME} & \textbf{TTFT} & \textbf{Throughput} \\
    & & & \textbf{(ms)} & \textbf{(tok/s)} \\
    \midrule
    OFF  & 64.2 & 66.4 & 3436 & 51 \\
    50\% & 63.7 & 66.5 & 2384 & 80 \\
    60\% & 63.4 & 66.2 & 2223 & 85 \\
    70\% & 62.8 & 65.8 & 2000 & 95 \\
    75\% & 62.3 & 65.7 & 1840 & 103 \\
    80\% & 62.8 & 65.1 & 1717 & 103 \\
    90\% & 60.4 & 64.0 & 1567 & 132 \\
    \bottomrule
    \end{tabular}
    \end{adjustbox}
  \end{minipage}%
  \hfill
  \begin{minipage}[t]{0.48\textwidth}
    \vspace{0pt}
    \centering
    \includegraphics[width=\linewidth]{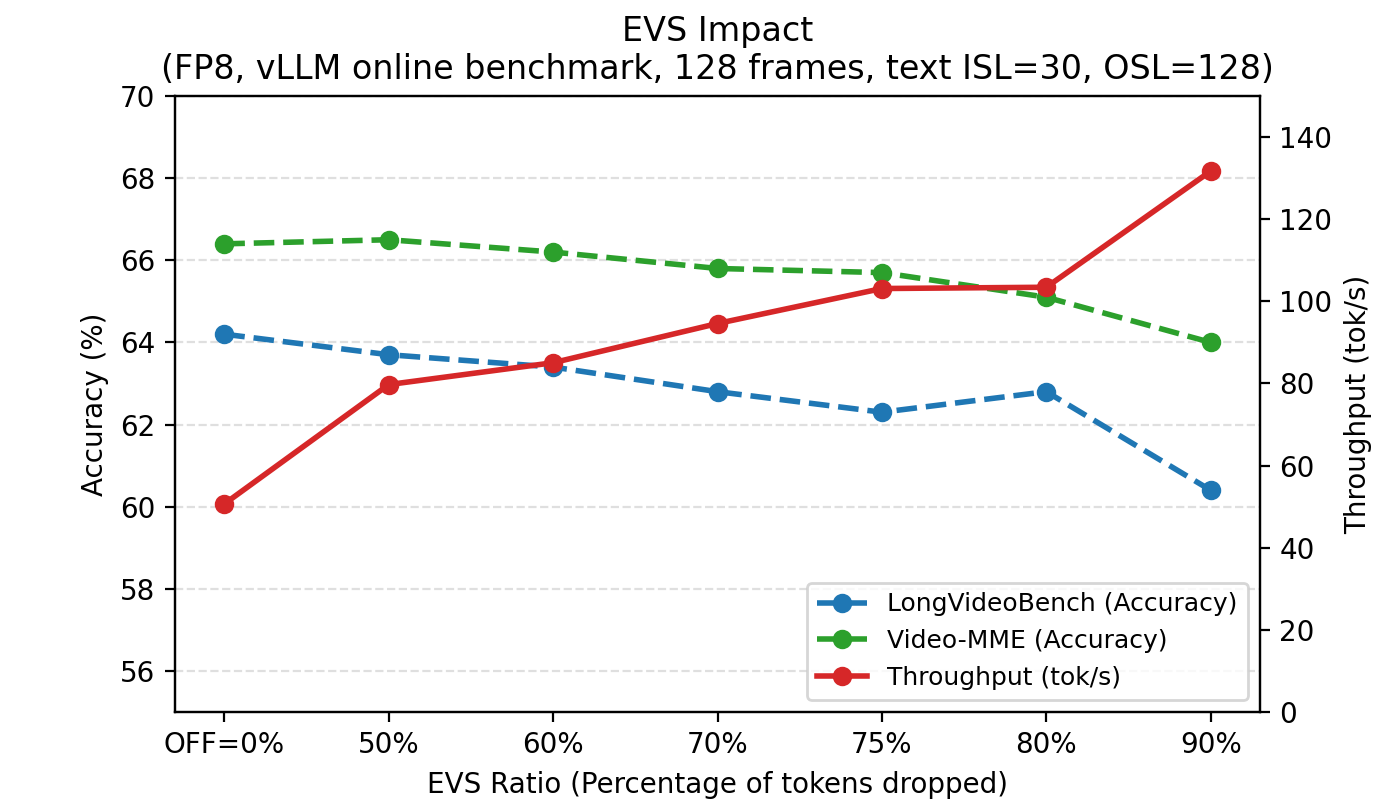}
  \end{minipage}

  \caption{EVS Ablation (RTX 6000 PRO SE, vLLM online benchmark, 128 frames, text ISL=30, OSL=128): top row shows BF16 results, bottom row shows FP8 results; left shows numeric tables (accuracy, time-TTFT, throughput), right shows corresponding visualizations.}
  \label{fig:evs_combo}
\end{figure*}

We evaluate the model on two video benchmarks: \textbf{Video-MME} and \textbf{LongVideoBench}.  
Figure~\ref{fig:evs_combo} presents EVS ablations for both BF16 and FP8 precision. The results demonstrate that EVS maintains strong performance on long-context video understanding tasks while significantly improving efficiency: as the EVS ratio increases, time-to-first-token (TTFT) decreases and throughput rises, with only a minor impact on accuracy.

\subsection{Image Processing Ablations}
\label{section:tiling_vs_dynamic_res}

We investigate an alternative approach to tiling for our input image processing pipeline. Instead of splitting the input image into non-overlapping tiles, we pass the entire image at native resolution to the vision encoder and MLP projector followed by convolutional token reduction for $4\times$ sequence compression.

For each image processing variant, we train the model through Stages 0 and 1 of our full training recipe, excluding the text-only reasoning datasets from Stage 1. Table \ref{tab:dynamic-res} compares these image processing strategies across a subset of multimodal benchmarks.

As shown in Columns 1 and 2, the native-resolution approach achieves comparable or even superior accuracy on several benchmarks. However, we observe a notable performance drop on OCRBench and OCRBench-V2 (English). Upon analysis, we found that the tiling algorithm occasionally applies large rescaling factors to smaller images to preserve aspect ratio. To isolate this effect, we conduct an additional experiment where we resize each image to match the size and aspect ratio that the tiling algorithm would have selected, without applying tiling. The results of this configuration are shown in Column 3 of Table \ref{tab:dynamic-res}, where we recover the performance drop on OCRBench. The gap on OCRBench-V2 (English), however, persists. We plan to further investigate strategies to address this gap in future work.

\begin{table*}[h!]
\centering
\scriptsize
\begin{tabular}{l *{5}{c}}
\toprule
\multicolumn{1}{l}{\textbf{Benchmark}} &
\multicolumn{1}{c}{\textbf{\shortstack[c]{Tiling}}} &
\multicolumn{1}{c}{\textbf{\shortstack[c]{Native Resolution}}} &
\multicolumn{1}{c}{\textbf{\shortstack[c]{Native Resolution\\with\\tiling-size matching}}} \\
\midrule
\texttt{AI2D (Test)}                 & 87.1 & 86.4 & \textbf{87.8} \\
\texttt{ChartQA (Test)}              & 89.8 & 88.4 & \textbf{90.3} \\
\texttt{DocVQA (Val)}                & 94.5 & \textbf{95.1} & 94.9 \\
\texttt{InfoVQA (Val)}               & 80.2 & \textbf{80.7} & 79.0 \\
\texttt{MMMU (Val)}                  & \textbf{56.8} & 56.0 & 56.2 \\
\texttt{MathVista-Mini}              & 69.7 & \textbf{71.3} & 71.2 \\
\texttt{OCRBench}                    & 84.5 & 82.8 & \textbf{85.3} \\
\texttt{OCRBench-V2 (CN)}            & 40.5 & \textbf{45.3} & 42.7 \\
\texttt{OCRBench-V2 (EN)}            & \textbf{61.4} & 57.6 & 57.6 \\
\texttt{TextVQA (Val)}               & 85.6 & 84.6 & \textbf{86.2} \\
\texttt{Average}                     & 75.0 & 74.8 & \textbf{75.1} \\
\bottomrule
\end{tabular}
\caption{We compare our image tiling strategy with: 1) Native resolution image inputs to the vision encoder followed by convolutional token reduction 2) Native resolution image inputs with tiling-size matching where we resize the image to the size and aspect ratio the dynamic tiling algorithm would have picked, and then feed the resized image to the vision encoder followed by convolutional token merging. We ran the benchmarks on the last 10 saved checkpoints and, for each strategy, select the one with the highest average benchmark score.}
\label{tab:dynamic-res}
\end{table*}

\subsection{Quantization}
\label{section:quantization}

The model is trained with Transformer Engine’s delayed-scaling FP8, using dynamic scaling factors computed from a running history of activation maxima across training iterations. By contrast, most downstream inference stacks (e.g.\ vLLM, TensorRT-LLM) assume static quantization with scaling factors fixed during a one-time calibration. To bridge this train–serve gap, we produce deployable checkpoints via post-training quantization (PTQ) and quantization-aware distillation (QAD) using software libraries TensorRT-Model-Optimizer and Megatron-LM. Unless otherwise noted, we quantize the language backbone - i.e., all linear layers, including both weights and activations - and retain the embedding layers, KV cache or others in higher precision. QAD is performed in BF16 to simulate lower-precision behavior. For FP8 we adopt the E4M3 format; we additionally provide NVFP4 variants from QAD. 

\paragraph{PTQ}
We calibrate on 1{,}024 samples drawn from the training set and compute per-tensor static scales for weights and activations using the absolute $\operatorname{amax}$ aggregated over the calibration dataset. The resulting FP8/NVFP4 checkpoints are directly usable in standard inference stacks such as vLLM.

\paragraph{QAD}
Because NVFP4 PTQ yields a small accuracy drop, we further improve quality by distilling a PTQ student from a BF16 teacher using a logit-matching loss (e.g.\ KL divergence) applied to the final model outputs only. Hyperparameters mirror SFT Stage~1, except we use a learning rate of $2\times10^{-6}$.

Table \ref{tab:vllm_precision_results} summarizes the accuracy comparison across different quantization precisions and strategies. All evaluations are conducted within the vLLM framework under BF16, FP8, and NVFP4 runtime configurations. Minor discrepancies relative to previously reported baselines are attributable to variations in evaluation framework implementations.

\begin{table}[h!]
\centering
\begin{tabular}{lccccc}
\toprule
\textbf{Precision} & \textbf{AI2D} & \textbf{ChartQA} & \textbf{OCRBench} & \textbf{DocVQA-val} & \textbf{OCRBenchV2 English} \\
\midrule
BF16      & 87.21 & 89.68 & 854 & 94.22 & 61.74 \\
FP8-PTQ   & 87.56 & 89.44 & 854 & 94.32 & 61.83 \\
NVFP4-PTQ   & 86.37 & 88.84 & 863 & 92.38 & 60.88 \\
NVFP4-QAD   & 87.14 & 89.96 & 851 & 93.95 & 61.94 \\
\bottomrule
\end{tabular}
\caption{Accuracy comparison across different precisions and tasks using vLLM backend.}
\label{tab:vllm_precision_results}
\end{table}

\section{Conclusion}

In this work, we introduced Nemotron Nano V2 VL, an efficient 12B vision–language model built upon the Nemotron-Nano-V2 LLM. The model demonstrates substantial improvements in multimodal and text understanding, as well as reasoning capabilities, compared to its predecessor, Llama-3.1-Nemotron-Nano-VL-8B. We employed a multi-stage training strategy to enhance visual understanding while preserving the text comprehension abilities of the original backbone. We presented a comprehensive evaluation of the model across diverse tasks and modalities, along with investigations into efficient video sampling at inference and alternative image processing pipelines. Finally, we have open-sourced the model weights in BF16, FP8, and FP4 formats, along with a significant portion of our SFT dataset and associated tooling (see Section \ref{section:introduction}).
\clearpage 
\clearpage 
\section{Contributors}

\textbf{Core Model Training and Data Study}

Amala Sanjay Deshmukh, Kateryna Chumachenko, Tuomas Rintamaki, Matthieu Le, Tyler Poon, Danial Mohseni Taheri, Ilia Karmanov, Guilin Liu, Jarno Seppanen, Guo Chen, Karan Sapra, Zhiding Yu, Adi Renduchintala, Charles Wang, Peter Jin.

\textbf{Systems, Data and Infrastructure }
Arushi Goel, Mike Ranzinger, Lukas Voegtle, Philipp Fischer, Timo Roman, Wei Ping, Boxin Wang, Zhuolin Yang, Nayeon Lee, Shaokun Zhang, Fuxiao Liu, Zhiqi Li, Di Zhang, Greg Heinrich, Hongxu (Danny) Yin, Song Han, Pavlo Molchanov, Parth Mannan, Yao Xu, Jane Polak Scowcroft, Tom Balough, Subhashree Radhakrishnan, Paris Zhang, Sean Cha, Ratnesh Kumar, Zaid Pervaiz Bhat, Jian Zhang, Darragh Hanley, Pritam Biswas, Jesse Oliver, Kevin Vasques, Roger Waleffe, Duncan Riach,  Oluwatobi Olabiyi, Ameya Sunil Mahabaleshwarkar, Bilal Kartal, Pritam Gundecha, Khanh Nguyen.

\textbf{Inference and Optimization}
Alexandre Milesi, Eugene Khvedchenia, Ran Zilberstein, Ofri Masad, Natan Bagrov, Nave Assaf, Tomer Asida, Daniel Afrimi, Amit Zuker, Netanel Haber, Zhiyu Cheng, Jingyu (Justin) Xin, Di (Allan) Wu, Nik Spirin, Maryam Moosaei, Roman Ageev, Vanshil Atul Shah, Yuting Wu, Daniel Korzekwa, Unnikrishnan Kizhakkemadam Sreekumar, Wanli Jiang.

\textbf{Evaluation, Safety, Product and Legal }
Padmavathy Subramanian, Alejandra Rico, Sandip Bhaskar, Saeid Motiian, Kedi Wu, Annie Surla, Chia-Chih Chen, Hayden Wolff, Matthew Feinberg, Melissa Corpuz, Marek Wawrzos, Eileen Long, Aastha Jhunjhunwala, Paul Hendricks, Farzan Memarian, Benika Hall, Xin-Yu Wang, David Mosallanezhad, Soumye Singhal, Luis Vega, Katherine Cheung, Krzysztof Pawelec, Michael Evans, Katherine Luna, Jie Lou, Erick Galinkin, Akshay Hazare, Kaustubh Purandare, Ann Guan, Anna Warno, Chen Cui, Yoshi Suhara, Shibani Likhite, Seph Mard, Meredith Price, Laya Sleiman, Saori Kaji.

\textbf{Leadership}
Udi Karpas, Kari Briski, Joey Conway, Michael Lightstone, Jan Kautz, Mohammad Shoeybi, Mostofa Patwary,  Jonathen Cohen, Oleksii Kuchaiev, Andrew Tao, Bryan Catanzaro.

\bibliography{references}

@misc{nvidia2025nvidianemotronnano2,
      title={NVIDIA Nemotron Nano 2: An Accurate and Efficient Hybrid Mamba-Transformer Reasoning Model}, 
      author={NVIDIA and : and Aarti Basant and Abhijit Khairnar and Abhijit Paithankar and Abhinav Khattar and Adithya Renduchintala and Aditya Malte and Akhiad Bercovich and Akshay Hazare and Alejandra Rico and Aleksander Ficek and Alex Kondratenko and Alex Shaposhnikov and Alexander Bukharin and Ali Taghibakhshi and Amelia Barton and Ameya Sunil Mahabaleshwarkar and Amy Shen and Andrew Tao and Ann Guan and Anna Shors and Anubhav Mandarwal and Arham Mehta and Arun Venkatesan and Ashton Sharabiani and Ashwath Aithal and Ashwin Poojary and Ayush Dattagupta and Balaram Buddharaju and Banghua Zhu and Barnaby Simkin and Bilal Kartal and Bita Darvish Rouhani and Bobby Chen and Boris Ginsburg and Brandon Norick and Brian Yu and Bryan Catanzaro and Charles Wang and Charlie Truong and Chetan Mungekar and Chintan Patel and Chris Alexiuk and Christian Munley and Christopher Parisien and Dan Su and Daniel Afrimi and Daniel Korzekwa and Daniel Rohrer and Daria Gitman and David Mosallanezhad and Deepak Narayanan and Dima Rekesh and Dina Yared and Dmytro Pykhtar and Dong Ahn and Duncan Riach and Eileen Long and Elliott Ning and Eric Chung and Erick Galinkin and Evelina Bakhturina and Gargi Prasad and Gerald Shen and Haifeng Qian and Haim Elisha and Harsh Sharma and Hayley Ross and Helen Ngo and Herman Sahota and Hexin Wang and Hoo Chang Shin and Hua Huang and Iain Cunningham and Igor Gitman and Ivan Moshkov and Jaehun Jung and Jan Kautz and Jane Polak Scowcroft and Jared Casper and Jian Zhang and Jiaqi Zeng and Jimmy Zhang and Jinze Xue and Jocelyn Huang and Joey Conway and John Kamalu and Jonathan Cohen and Joseph Jennings and Julien Veron Vialard and Junkeun Yi and Jupinder Parmar and Kari Briski and Katherine Cheung and Katherine Luna and Keith Wyss and Keshav Santhanam and Kezhi Kong and Krzysztof Pawelec and Kumar Anik and Kunlun Li and Kushan Ahmadian and Lawrence McAfee and Laya Sleiman and Leon Derczynski and Luis Vega and Maer Rodrigues de Melo and Makesh Narsimhan Sreedhar and Marcin Chochowski and Mark Cai and Markus Kliegl and Marta Stepniewska-Dziubinska and Matvei Novikov and Mehrzad Samadi and Meredith Price and Meriem Boubdir and Michael Boone and Michael Evans and Michal Bien and Michal Zawalski and Miguel Martinez and Mike Chrzanowski and Mohammad Shoeybi and Mostofa Patwary and Namit Dhameja and Nave Assaf and Negar Habibi and Nidhi Bhatia and Nikki Pope and Nima Tajbakhsh and Nirmal Kumar Juluru and Oleg Rybakov and Oleksii Hrinchuk and Oleksii Kuchaiev and Oluwatobi Olabiyi and Pablo Ribalta and Padmavathy Subramanian and Parth Chadha and Pavlo Molchanov and Peter Dykas and Peter Jin and Piotr Bialecki and Piotr Januszewski and Pradeep Thalasta and Prashant Gaikwad and Prasoon Varshney and Pritam Gundecha and Przemek Tredak and Rabeeh Karimi Mahabadi and Rajen Patel and Ran El-Yaniv and Ranjit Rajan and Ria Cheruvu and Rima Shahbazyan and Ritika Borkar and Ritu Gala and Roger Waleffe and Ruoxi Zhang and Russell J. Hewett and Ryan Prenger and Sahil Jain and Samuel Kriman and Sanjeev Satheesh and Saori Kaji and Sarah Yurick and Saurav Muralidharan and Sean Narenthiran and Seonmyeong Bak and Sepehr Sameni and Seungju Han and Shanmugam Ramasamy and Shaona Ghosh and Sharath Turuvekere Sreenivas and Shelby Thomas and Shizhe Diao and Shreya Gopal and Shrimai Prabhumoye and Shubham Toshniwal and Shuoyang Ding and Siddharth Singh and Siddhartha Jain and Somshubra Majumdar and Soumye Singhal and Stefania Alborghetti and Syeda Nahida Akter and Terry Kong and Tim Moon and Tomasz Hliwiak and Tomer Asida and Tony Wang and Tugrul Konuk and Twinkle Vashishth and Tyler Poon and Udi Karpas and Vahid Noroozi and Venkat Srinivasan and Vijay Korthikanti and Vikram Fugro and Vineeth Kalluru and Vitaly Kurin and Vitaly Lavrukhin and Wasi Uddin Ahmad and Wei Du and Wonmin Byeon and Ximing Lu and Xin Dong and Yashaswi Karnati and Yejin Choi and Yian Zhang and Ying Lin and Yonggan Fu and Yoshi Suhara and Zhen Dong and Zhiyu Li and Zhongbo Zhu and Zijia Chen},
      year={2025},
      eprint={2508.14444},
      archivePrefix={arXiv},
      primaryClass={cs.CL},
      url={https://arxiv.org/abs/2508.14444}, 
}

@misc{liu2024improvedbaselinesvisualinstruction,
      title={Improved Baselines with Visual Instruction Tuning}, 
      author={Haotian Liu and Chunyuan Li and Yuheng Li and Yong Jae Lee},
      year={2024},
      eprint={2310.03744},
      archivePrefix={arXiv},
      primaryClass={cs.CV},
      url={https://arxiv.org/abs/2310.03744}, 
}

@misc{li2025eagle2buildingposttraining,
      title={Eagle 2: Building Post-Training Data Strategies from Scratch for Frontier Vision-Language Models}, 
      author={Zhiqi Li and Guo Chen and Shilong Liu and Shihao Wang and Vibashan VS and Yishen Ji and Shiyi Lan and Hao Zhang and Yilin Zhao and Subhashree Radhakrishnan and Nadine Chang and Karan Sapra and Amala Sanjay Deshmukh and Tuomas Rintamaki and Matthieu Le and Ilia Karmanov and Lukas Voegtle and Philipp Fischer and De-An Huang and Timo Roman and Tong Lu and Jose M. Alvarez and Bryan Catanzaro and Jan Kautz and Andrew Tao and Guilin Liu and Zhiding Yu},
      year={2025},
      eprint={2501.14818},
      archivePrefix={arXiv},
      primaryClass={cs.CV},
      url={https://arxiv.org/abs/2501.14818}, 
}

@misc{chen2025eagle25boostinglongcontext,
      title={Eagle 2.5: Boosting Long-Context Post-Training for Frontier Vision-Language Models}, 
      author={Guo Chen and Zhiqi Li and Shihao Wang and Jindong Jiang and Yicheng Liu and Lidong Lu and De-An Huang and Wonmin Byeon and Matthieu Le and Tuomas Rintamaki and Tyler Poon and Max Ehrlich and Tuomas Rintamaki and Tyler Poon and Tong Lu and Limin Wang and Bryan Catanzaro and Jan Kautz and Andrew Tao and Zhiding Yu and Guilin Liu},
      year={2025},
      eprint={2504.15271},
      archivePrefix={arXiv},
      primaryClass={cs.CV},
      url={https://arxiv.org/abs/2504.15271}, 
}

@misc{chen2024fargpt4vclosinggap,
      title={How Far Are We to GPT-4V? Closing the Gap to Commercial Multimodal Models with Open-Source Suites}, 
      author={Zhe Chen and Weiyun Wang and Hao Tian and Shenglong Ye and Zhangwei Gao and Erfei Cui and Wenwen Tong and Kongzhi Hu and Jiapeng Luo and Zheng Ma and Ji Ma and Jiaqi Wang and Xiaoyi Dong and Hang Yan and Hewei Guo and Conghui He and Botian Shi and Zhenjiang Jin and Chao Xu and Bin Wang and Xingjian Wei and Wei Li and Wenjian Zhang and Bo Zhang and Pinlong Cai and Licheng Wen and Xiangchao Yan and Min Dou and Lewei Lu and Xizhou Zhu and Tong Lu and Dahua Lin and Yu Qiao and Jifeng Dai and Wenhai Wang},
      year={2024},
      eprint={2404.16821},
      archivePrefix={arXiv},
      primaryClass={cs.CV},
      url={https://arxiv.org/abs/2404.16821}, 
}

@misc{chen2025expandingperformanceboundariesopensource,
      title={Expanding Performance Boundaries of Open-Source Multimodal Models with Model, Data, and Test-Time Scaling}, 
      author={Zhe Chen and Weiyun Wang and Yue Cao and Yangzhou Liu and Zhangwei Gao and Erfei Cui and Jinguo Zhu and Shenglong Ye and Hao Tian and Zhaoyang Liu and Lixin Gu and Xuehui Wang and Qingyun Li and Yiming Ren and Zixuan Chen and Jiapeng Luo and Jiahao Wang and Tan Jiang and Bo Wang and Conghui He and Botian Shi and Xingcheng Zhang and Han Lv and Yi Wang and Wenqi Shao and Pei Chu and Zhongying Tu and Tong He and Zhiyong Wu and Huipeng Deng and Jiaye Ge and Kai Chen and Kaipeng Zhang and Limin Wang and Min Dou and Lewei Lu and Xizhou Zhu and Tong Lu and Dahua Lin and Yu Qiao and Jifeng Dai and Wenhai Wang},
      year={2025},
      eprint={2412.05271},
      archivePrefix={arXiv},
      primaryClass={cs.CV},
      url={https://arxiv.org/abs/2412.05271}, 
}

@article{megatron-lm,
  title={Megatron-LM: Training Multi-Billion Parameter Language Models Using Model Parallelism},
  author={Shoeybi, Mohammad and Patwary, Mostofa and Puri, Raul and LeGresley, Patrick and Casper, Jared and Catanzaro, Bryan},
  journal={arXiv preprint arXiv:1909.08053},
  year={2019}
}

@misc{duan2025vlmevalkitopensourcetoolkitevaluating,
      title={VLMEvalKit: An Open-Source Toolkit for Evaluating Large Multi-Modality Models}, 
      author={Haodong Duan and Xinyu Fang and Junming Yang and Xiangyu Zhao and Yuxuan Qiao and Mo Li and Amit Agarwal and Zhe Chen and Lin Chen and Yuan Liu and Yubo Ma and Hailong Sun and Yifan Zhang and Shiyin Lu and Tack Hwa Wong and Weiyun Wang and Peiheng Zhou and Xiaozhe Li and Chaoyou Fu and Junbo Cui and Jixuan Chen and Enxin Song and Song Mao and Shengyuan Ding and Tianhao Liang and Zicheng Zhang and Xiaoyi Dong and Yuhang Zang and Pan Zhang and Jiaqi Wang and Dahua Lin and Kai Chen},
      year={2025},
      eprint={2407.11691},
      archivePrefix={arXiv},
      primaryClass={cs.CV},
      url={https://arxiv.org/abs/2407.11691}, 
}

@inproceedings{kwon2023efficient,
  title={Efficient Memory Management for Large Language Model Serving with PagedAttention},
  author={Woosuk Kwon and Zhuohan Li and Siyuan Zhuang and Ying Sheng and Lianmin Zheng and Cody Hao Yu and Joseph E. Gonzalez and Hao Zhang and Ion Stoica},
  booktitle={Proceedings of the ACM SIGOPS 29th Symposium on Operating Systems Principles},
  year={2023}
}

@misc{lightman2023letsverifystepstep,
      title={Let's Verify Step by Step}, 
      author={Hunter Lightman and Vineet Kosaraju and Yura Burda and Harri Edwards and Bowen Baker and Teddy Lee and Jan Leike and John Schulman and Ilya Sutskever and Karl Cobbe},
      year={2023},
      eprint={2305.20050},
      archivePrefix={arXiv},
      primaryClass={cs.LG},
      url={https://arxiv.org/abs/2305.20050}, 
}

@misc{heinrich2025radiov25improvedbaselinesagglomerative,
      title={RADIOv2.5: Improved Baselines for Agglomerative Vision Foundation Models}, 
      author={Greg Heinrich and Mike Ranzinger and Hongxu and Yin and Yao Lu and Jan Kautz and Andrew Tao and Bryan Catanzaro and Pavlo Molchanov},
      year={2025},
      eprint={2412.07679},
      archivePrefix={arXiv},
      primaryClass={cs.CV},
      url={https://arxiv.org/abs/2412.07679}, 
}

@misc{rein2023gpqagraduatelevelgoogleproofqa,
      title={GPQA: A Graduate-Level Google-Proof Q\&A Benchmark}, 
      author={David Rein and Betty Li Hou and Asa Cooper Stickland and Jackson Petty and Richard Yuanzhe Pang and Julien Dirani and Julian Michael and Samuel R. Bowman},
      year={2023},
      eprint={2311.12022},
      archivePrefix={arXiv},
      primaryClass={cs.AI},
      url={https://arxiv.org/abs/2311.12022}, 
}

@misc{jain2024livecodebenchholisticcontaminationfree,
      title={LiveCodeBench: Holistic and Contamination Free Evaluation of Large Language Models for Code}, 
      author={Naman Jain and King Han and Alex Gu and Wen-Ding Li and Fanjia Yan and Tianjun Zhang and Sida Wang and Armando Solar-Lezama and Koushik Sen and Ion Stoica},
      year={2024},
      eprint={2403.07974},
      archivePrefix={arXiv},
      primaryClass={cs.SE},
      url={https://arxiv.org/abs/2403.07974}, 
}

@misc{zhou2023instructionfollowingevaluationlargelanguage,
      title={Instruction-Following Evaluation for Large Language Models}, 
      author={Jeffrey Zhou and Tianjian Lu and Swaroop Mishra and Siddhartha Brahma and Sujoy Basu and Yi Luan and Denny Zhou and Le Hou},
      year={2023},
      eprint={2311.07911},
      archivePrefix={arXiv},
      primaryClass={cs.CL},
      url={https://arxiv.org/abs/2311.07911}, 
}

@misc{tian2024scicoderesearchcodingbenchmark,
      title={SciCode: A Research Coding Benchmark Curated by Scientists}, 
      author={Minyang Tian and Luyu Gao and Shizhuo Dylan Zhang and Xinan Chen and Cunwei Fan and Xuefei Guo and Roland Haas and Pan Ji and Kittithat Krongchon and Yao Li and Shengyan Liu and Di Luo and Yutao Ma and Hao Tong and Kha Trinh and Chenyu Tian and Zihan Wang and Bohao Wu and Yanyu Xiong and Shengzhu Yin and Minhui Zhu and Kilian Lieret and Yanxin Lu and Genglin Liu and Yufeng Du and Tianhua Tao and Ofir Press and Jamie Callan and Eliu Huerta and Hao Peng},
      year={2024},
      eprint={2407.13168},
      archivePrefix={arXiv},
      primaryClass={cs.AI},
      url={https://arxiv.org/abs/2407.13168}, 
}

@misc{hsieh2024rulerwhatsrealcontext,
      title={RULER: What's the Real Context Size of Your Long-Context Language Models?}, 
      author={Cheng-Ping Hsieh and Simeng Sun and Samuel Kriman and Shantanu Acharya and Dima Rekesh and Fei Jia and Yang Zhang and Boris Ginsburg},
      year={2024},
      eprint={2404.06654},
      archivePrefix={arXiv},
      primaryClass={cs.CL},
      url={https://arxiv.org/abs/2404.06654}, 
}

@misc{wang2024mmluprorobustchallengingmultitask,
      title={MMLU-Pro: A More Robust and Challenging Multi-Task Language Understanding Benchmark}, 
      author={Yubo Wang and Xueguang Ma and Ge Zhang and Yuansheng Ni and Abhranil Chandra and Shiguang Guo and Weiming Ren and Aaran Arulraj and Xuan He and Ziyan Jiang and Tianle Li and Max Ku and Kai Wang and Alex Zhuang and Rongqi Fan and Xiang Yue and Wenhu Chen},
      year={2024},
      eprint={2406.01574},
      archivePrefix={arXiv},
      primaryClass={cs.CL},
      url={https://arxiv.org/abs/2406.01574}, 
}

@misc{wang2025internvl35advancingopensourcemultimodal,
      title={InternVL3.5: Advancing Open-Source Multimodal Models in Versatility, Reasoning, and Efficiency}, 
      author={Weiyun Wang and Zhangwei Gao and Lixin Gu and Hengjun Pu and Long Cui and Xingguang Wei and Zhaoyang Liu and Linglin Jing and Shenglong Ye and Jie Shao and Zhaokai Wang and Zhe Chen and Hongjie Zhang and Ganlin Yang and Haomin Wang and Qi Wei and Jinhui Yin and Wenhao Li and Erfei Cui and Guanzhou Chen and Zichen Ding and Changyao Tian and Zhenyu Wu and Jingjing Xie and Zehao Li and Bowen Yang and Yuchen Duan and Xuehui Wang and Zhi Hou and Haoran Hao and Tianyi Zhang and Songze Li and Xiangyu Zhao and Haodong Duan and Nianchen Deng and Bin Fu and Yinan He and Yi Wang and Conghui He and Botian Shi and Junjun He and Yingtong Xiong and Han Lv and Lijun Wu and Wenqi Shao and Kaipeng Zhang and Huipeng Deng and Biqing Qi and Jiaye Ge and Qipeng Guo and Wenwei Zhang and Songyang Zhang and Maosong Cao and Junyao Lin and Kexian Tang and Jianfei Gao and Haian Huang and Yuzhe Gu and Chengqi Lyu and Huanze Tang and Rui Wang and Haijun Lv and Wanli Ouyang and Limin Wang and Min Dou and Xizhou Zhu and Tong Lu and Dahua Lin and Jifeng Dai and Weijie Su and Bowen Zhou and Kai Chen and Yu Qiao and Wenhai Wang and Gen Luo},
      year={2025},
      eprint={2508.18265},
      archivePrefix={arXiv},
      primaryClass={cs.CV},
      url={https://arxiv.org/abs/2508.18265}, 
}

@misc{vteam2025glm45vglm41vthinkingversatilemultimodal,
      title={GLM-4.5V and GLM-4.1V-Thinking: Towards Versatile Multimodal Reasoning with Scalable Reinforcement Learning}, 
      author={Wenyi Hong and Wenmeng Yu and Xiaotao Gu and Guo Wang and Guobing Gan and Haomiao Tang and Jiale Cheng and Ji Qi and Junhui Ji and Lihang Pan and Shuaiqi Duan and Weihan Wang and Yan Wang and Yean Cheng and Zehai He and Zhe Su and Zhen Yang and Ziyang Pan and Aohan Zeng and Baoxu Wang and Bin Chen and Boyan Shi and Changyu Pang and Chenhui Zhang and Da Yin and Fan Yang and Guoqing Chen and Jiazheng Xu and Jiale Zhu and Jiali Chen and Jing Chen and Jinhao Chen and Jinghao Lin and Jinjiang Wang and Junjie Chen and Leqi Lei and Letian Gong and Leyi Pan and Mingdao Liu and Mingde Xu and Mingzhi Zhang and Qinkai Zheng and Sheng Yang and Shi Zhong and Shiyu Huang and Shuyuan Zhao and Siyan Xue and Shangqin Tu and Shengbiao Meng and Tianshu Zhang and Tianwei Luo and Tianxiang Hao and Tianyu Tong and Wenkai Li and Wei Jia and Xiao Liu and Xiaohan Zhang and Xin Lyu and Xinyue Fan and Xuancheng Huang and Yanling Wang and Yadong Xue and Yanfeng Wang and Yanzi Wang and Yifan An and Yifan Du and Yiming Shi and Yiheng Huang and Yilin Niu and Yuan Wang and Yuanchang Yue and Yuchen Li and Yutao Zhang and Yuting Wang and Yu Wang and Yuxuan Zhang and Zhao Xue and Zhenyu Hou and Zhengxiao Du and Zihan Wang and Peng Zhang and Debing Liu and Bin Xu and Juanzi Li and Minlie Huang and Yuxiao Dong and Jie Tang},
      year={2025},
      eprint={2507.01006},
      archivePrefix={arXiv},
      primaryClass={cs.CV},
      url={https://arxiv.org/abs/2507.01006}, 
}

@misc{yang2025qwen3technicalreport,
      title={Qwen3 Technical Report}, 
      author={An Yang and Anfeng Li and Baosong Yang and Beichen Zhang and Binyuan Hui and Bo Zheng and Bowen Yu and Chang Gao and Chengen Huang and Chenxu Lv and Chujie Zheng and Dayiheng Liu and Fan Zhou and Fei Huang and Feng Hu and Hao Ge and Haoran Wei and Huan Lin and Jialong Tang and Jian Yang and Jianhong Tu and Jianwei Zhang and Jianxin Yang and Jiaxi Yang and Jing Zhou and Jingren Zhou and Junyang Lin and Kai Dang and Keqin Bao and Kexin Yang and Le Yu and Lianghao Deng and Mei Li and Mingfeng Xue and Mingze Li and Pei Zhang and Peng Wang and Qin Zhu and Rui Men and Ruize Gao and Shixuan Liu and Shuang Luo and Tianhao Li and Tianyi Tang and Wenbiao Yin and Xingzhang Ren and Xinyu Wang and Xinyu Zhang and Xuancheng Ren and Yang Fan and Yang Su and Yichang Zhang and Yinger Zhang and Yu Wan and Yuqiong Liu and Zekun Wang and Zeyu Cui and Zhenru Zhang and Zhipeng Zhou and Zihan Qiu},
      year={2025},
      eprint={2505.09388},
      archivePrefix={arXiv},
      primaryClass={cs.CL},
      url={https://arxiv.org/abs/2505.09388}, 
}

@misc{bai2025qwen25vltechnicalreport,
      title={Qwen2.5-{VL} Technical Report}, 
      author={Shuai Bai and Keqin Chen and Xuejing Liu and Jialin Wang and Wenbin Ge and Sibo Song and Kai Dang and Peng Wang and Shijie Wang and Jun Tang and Humen Zhong and Yuanzhi Zhu and Mingkun Yang and Zhaohai Li and Jianqiang Wan and Pengfei Wang and Wei Ding and Zheren Fu and Yiheng Xu and Jiabo Ye and Xi Zhang and Tianbao Xie and Zesen Cheng and Hang Zhang and Zhibo Yang and Haiyang Xu and Junyang Lin},
      year={2025},
      eprint={2502.13923},
      archivePrefix={arXiv},
      primaryClass={cs.CV},
      url={https://arxiv.org/abs/2502.13923}, 
}

@misc{qwen2025qwen25technicalreport,
      title={Qwen2.5 Technical Report}, 
      author={Qwen and : and An Yang and Baosong Yang and Beichen Zhang and Binyuan Hui and Bo Zheng and Bowen Yu and Chengyuan Li and Dayiheng Liu and Fei Huang and Haoran Wei and Huan Lin and Jian Yang and Jianhong Tu and Jianwei Zhang and Jianxin Yang and Jiaxi Yang and Jingren Zhou and Junyang Lin and Kai Dang and Keming Lu and Keqin Bao and Kexin Yang and Le Yu and Mei Li and Mingfeng Xue and Pei Zhang and Qin Zhu and Rui Men and Runji Lin and Tianhao Li and Tianyi Tang and Tingyu Xia and Xingzhang Ren and Xuancheng Ren and Yang Fan and Yang Su and Yichang Zhang and Yu Wan and Yuqiong Liu and Zeyu Cui and Zhenru Zhang and Zihan Qiu},
      year={2025},
      eprint={2412.15115},
      archivePrefix={arXiv},
      primaryClass={cs.CL},
      url={https://arxiv.org/abs/2412.15115}, 
}

@misc{fu2025videommefirstevercomprehensiveevaluation,
      title={Video-MME: The First-Ever Comprehensive Evaluation Benchmark of Multi-modal LLMs in Video Analysis}, 
      author={Chaoyou Fu and Yuhan Dai and Yongdong Luo and Lei Li and Shuhuai Ren and Renrui Zhang and Zihan Wang and Chenyu Zhou and Yunhang Shen and Mengdan Zhang and Peixian Chen and Yanwei Li and Shaohui Lin and Sirui Zhao and Ke Li and Tong Xu and Xiawu Zheng and Enhong Chen and Caifeng Shan and Ran He and Xing Sun},
      year={2025},
      eprint={2405.21075},
      archivePrefix={arXiv},
      primaryClass={cs.CV},
      url={https://arxiv.org/abs/2405.21075}, 
}

@misc{ma2024mmlongbenchdocbenchmarkinglongcontextdocument,
      title={MMLongBench-Doc: Benchmarking Long-context Document Understanding with Visualizations}, 
      author={Yubo Ma and Yuhang Zang and Liangyu Chen and Meiqi Chen and Yizhu Jiao and Xinze Li and Xinyuan Lu and Ziyu Liu and Yan Ma and Xiaoyi Dong and Pan Zhang and Liangming Pan and Yu-Gang Jiang and Jiaqi Wang and Yixin Cao and Aixin Sun},
      year={2024},
      eprint={2407.01523},
      archivePrefix={arXiv},
      primaryClass={cs.CV},
      url={https://arxiv.org/abs/2407.01523}, 
}

@misc{MegatronContextParallelism,
  title        = {{context\_ parallel package}},
  author       = {{Megatron Core}},
  year         = {2025},
  month        = {September},
  day          = {25},
  url          = {https://docs.nvidia.com/megatron-core/developer-guide/latest/api-guide/context_parallel.html}
}

@misc{chen2024longvilascalinglongcontextvisual,
      title={LongVILA: Scaling Long-Context Visual Language Models for Long Videos}, 
      author={Yukang Chen and Fuzhao Xue and Dacheng Li and Qinghao Hu and Ligeng Zhu and Xiuyu Li and Yunhao Fang and Haotian Tang and Shang Yang and Zhijian Liu and Ethan He and Hongxu Yin and Pavlo Molchanov and Jan Kautz and Linxi Fan and Yuke Zhu and Yao Lu and Song Han},
      year={2024},
      eprint={2408.10188},
      archivePrefix={arXiv},
      primaryClass={cs.CV},
      url={https://arxiv.org/abs/2408.10188}, 
}

@misc{tang2025mtvqabenchmarkingmultilingualtextcentric,
      title={MTVQA: Benchmarking Multilingual Text-Centric Visual Question Answering}, 
      author={Jingqun Tang and Qi Liu and Yongjie Ye and Jinghui Lu and Shu Wei and Chunhui Lin and Wanqing Li and Mohamad Fitri Faiz Bin Mahmood and Hao Feng and Zhen Zhao and Yangfan He and Kuan Lu and Yanjie Wang and Yuliang Liu and Hao Liu and Xiang Bai and Can Huang},
      year={2025},
      eprint={2405.11985},
      archivePrefix={arXiv},
      primaryClass={cs.CV},
      url={https://arxiv.org/abs/2405.11985}, 
}

@misc{sun2025parrotmultilingualvisualinstruction,
      title={Parrot: Multilingual Visual Instruction Tuning}, 
      author={Hai-Long Sun and Da-Wei Zhou and Yang Li and Shiyin Lu and Chao Yi and Qing-Guo Chen and Zhao Xu and Weihua Luo and Kaifu Zhang and De-Chuan Zhan and Han-Jia Ye},
      year={2025},
      eprint={2406.02539},
      archivePrefix={arXiv},
      primaryClass={cs.CV},
      url={https://arxiv.org/abs/2406.02539}, 
}

@misc{bagrov2025efficientvideosamplingpruning,
      title={Efficient Video Sampling: Pruning Temporally Redundant Tokens for Faster {VLM} Inference}, 
      author={Natan Bagrov and Eugene Khvedchenia and Borys Tymchenko and Shay Aharon and Lior Kadoch and Tomer Keren and Ofri Masad and Yonatan Geifman and Ran Zilberstein and Tuomas Rintamaki and Matthieu Le and Andrew Tao},
      year={2025},
      eprint={2510.14624},
      archivePrefix={arXiv},
      primaryClass={cs.CV},
      url={https://arxiv.org/abs/2510.14624}, 
}

@misc{liu2024mmbenchmultimodalmodelallaround,
      title={MMBench: Is Your Multi-modal Model an All-around Player?}, 
      author={Yuan Liu and Haodong Duan and Yuanhan Zhang and Bo Li and Songyang Zhang and Wangbo Zhao and Yike Yuan and Jiaqi Wang and Conghui He and Ziwei Liu and Kai Chen and Dahua Lin},
      year={2024},
      eprint={2307.06281},
      archivePrefix={arXiv},
      primaryClass={cs.CV},
      url={https://arxiv.org/abs/2307.06281}, 
}

@article{Liu_2024,
   title={OCRBench: on the hidden mystery of OCR in large multimodal models},
   volume={67},
   ISSN={1869-1919},
   url={http://dx.doi.org/10.1007/s11432-024-4235-6},
   DOI={10.1007/s11432-024-4235-6},
   number={12},
   journal={Science China Information Sciences},
   publisher={Springer Science and Business Media LLC},
   author={Liu, Yuliang and Li, Zhang and Huang, Mingxin and Yang, Biao and Yu, Wenwen and Li, Chunyuan and Yin, Xu-Cheng and Liu, Cheng-Lin and Jin, Lianwen and Bai, Xiang},
   year={2024},
   month=dec }

@misc{fu2024ocrbenchv2improvedbenchmark,
    title={OCRBench v2: An Improved Benchmark for Evaluating Large Multimodal Models on Visual Text Localization and Reasoning}, 
    author={Ling Fu and Zhebin Kuang and Jiajun Song and Mingxin Huang and Biao Yang and Yuzhe Li and Linghao Zhu and Qidi Luo and Xinyu Wang and Hao Lu and Zhang Li and Guozhi Tang and Bin Shan and Chunhui Lin and Qi Liu and Binghong Wu and Hao Feng and Hao Liu and Can Huang and Jingqun Tang and Wei Chen and Lianwen Jin and Yuliang Liu and Xiang Bai},
    year={2024},
    eprint={2501.00321},
    archivePrefix={arXiv},
    primaryClass={cs.CV},
    url={https://arxiv.org/abs/2501.00321}, 
}

@misc{carreira2018quovadisactionrecognition,
      title={Quo Vadis, Action Recognition? {A} New Model and the Kinetics Dataset}, 
      author={Joao Carreira and Andrew Zisserman},
      year={2018},
      eprint={1705.07750},
      archivePrefix={arXiv},
      primaryClass={cs.CV},
      url={https://arxiv.org/abs/1705.07750}, 
}

@misc{zhou2017automaticlearningproceduresweb,
      title={Towards Automatic Learning of Procedures from Web Instructional Videos}, 
      author={Luowei Zhou and Chenliang Xu and Jason J. Corso},
      year={2017},
      eprint={1703.09788},
      archivePrefix={arXiv},
      primaryClass={cs.CV},
      url={https://arxiv.org/abs/1703.09788}, 
}

@misc{zala2023hierarchicalvideomomentretrievalstepcaptioning,
      title={Hierarchical Video-Moment Retrieval and Step-Captioning}, 
      author={Abhay Zala and Jaemin Cho and Satwik Kottur and Xilun Chen and Barlas Oğuz and Yasher Mehdad and Mohit Bansal},
      year={2023},
      eprint={2303.16406},
      archivePrefix={arXiv},
      primaryClass={cs.CV},
      url={https://arxiv.org/abs/2303.16406}, 
}

@INPROCEEDINGS{7298698,
  author={Heilbron, Fabian Caba and Escorcia, Victor and Ghanem, Bernard and Niebles, Juan Carlos},
  booktitle={2015 IEEE Conference on Computer Vision and Pattern Recognition (CVPR)}, 
  title={ActivityNet: A large-scale video benchmark for human activity understanding}, 
  year={2015},
  volume={},
  number={},
  pages={961-970},
  doi={10.1109/CVPR.2015.7298698}}

@misc{huang2025egoexolearndatasetbridgingasynchronous,
      title={EgoExoLearn: A Dataset for Bridging Asynchronous Ego- and Exo-centric View of Procedural Activities in Real World}, 
      author={Yifei Huang and Guo Chen and Jilan Xu and Mingfang Zhang and Lijin Yang and Baoqi Pei and Hongjie Zhang and Lu Dong and Yali Wang and Limin Wang and Yu Qiao},
      year={2025},
      eprint={2403.16182},
      archivePrefix={arXiv},
      primaryClass={cs.CV},
      url={https://arxiv.org/abs/2403.16182}, 
}

@INPROCEEDINGS{6909500,
  author={Kuehne, Hilde and Arslan, Ali and Serre, Thomas},
  booktitle={2014 IEEE Conference on Computer Vision and Pattern Recognition}, 
  title={The Language of Actions: Recovering the Syntax and Semantics of Goal-Directed Human Activities}, 
  year={2014},
  volume={},
  number={},
  pages={780-787},
  doi={10.1109/CVPR.2014.105}}

@misc{pătrăucean2023perceptiontestdiagnosticbenchmark,
      title={Perception Test: A Diagnostic Benchmark for Multimodal Video Models}, 
      author={Viorica Pătrăucean and Lucas Smaira and Ankush Gupta and Adrià Recasens Continente and Larisa Markeeva and Dylan Banarse and Skanda Koppula and Joseph Heyward and Mateusz Malinowski and Yi Yang and Carl Doersch and Tatiana Matejovicova and Yury Sulsky and Antoine Miech and Alex Frechette and Hanna Klimczak and Raphael Koster and Junlin Zhang and Stephanie Winkler and Yusuf Aytar and Simon Osindero and Dima Damen and Andrew Zisserman and João Carreira},
      year={2023},
      eprint={2305.13786},
      archivePrefix={arXiv},
      primaryClass={cs.CV},
      url={https://arxiv.org/abs/2305.13786}, 
}

@misc{zhao2019hacshumanactionclips,
      title={HACS: Human Action Clips and Segments Dataset for Recognition and Temporal Localization}, 
      author={Hang Zhao and Antonio Torralba and Lorenzo Torresani and Zhicheng Yan},
      year={2019},
      eprint={1712.09374},
      archivePrefix={arXiv},
      primaryClass={cs.CV},
      url={https://arxiv.org/abs/1712.09374}, 
}

@misc{liu2022fineactionfinegrainedvideodataset,
      title={FineAction: A Fine-Grained Video Dataset for Temporal Action Localization}, 
      author={Yi Liu and Limin Wang and Yali Wang and Xiao Ma and Yu Qiao},
      year={2022},
      eprint={2105.11107},
      archivePrefix={arXiv},
      primaryClass={cs.CV},
      url={https://arxiv.org/abs/2105.11107}, 
}

@misc{grauman2022ego4dworld3000hours,
      title={Ego4D: Around the World in 3,000 Hours of Egocentric Video}, 
      author={Kristen Grauman and Andrew Westbury and Eugene Byrne and Zachary Chavis and Antonino Furnari and Rohit Girdhar and Jackson Hamburger and Hao Jiang and Miao Liu and Xingyu Liu and Miguel Martin and Tushar Nagarajan and Ilija Radosavovic and Santhosh Kumar Ramakrishnan and Fiona Ryan and Jayant Sharma and Michael Wray and Mengmeng Xu and Eric Zhongcong Xu and Chen Zhao and Siddhant Bansal and Dhruv Batra and Vincent Cartillier and Sean Crane and Tien Do and Morrie Doulaty and Akshay Erapalli and Christoph Feichtenhofer and Adriano Fragomeni and Qichen Fu and Abrham Gebreselasie and Cristina Gonzalez and James Hillis and Xuhua Huang and Yifei Huang and Wenqi Jia and Weslie Khoo and Jachym Kolar and Satwik Kottur and Anurag Kumar and Federico Landini and Chao Li and Yanghao Li and Zhenqiang Li and Karttikeya Mangalam and Raghava Modhugu and Jonathan Munro and Tullie Murrell and Takumi Nishiyasu and Will Price and Paola Ruiz Puentes and Merey Ramazanova and Leda Sari and Kiran Somasundaram and Audrey Southerland and Yusuke Sugano and Ruijie Tao and Minh Vo and Yuchen Wang and Xindi Wu and Takuma Yagi and Ziwei Zhao and Yunyi Zhu and Pablo Arbelaez and David Crandall and Dima Damen and Giovanni Maria Farinella and Christian Fuegen and Bernard Ghanem and Vamsi Krishna Ithapu and C. V. Jawahar and Hanbyul Joo and Kris Kitani and Haizhou Li and Richard Newcombe and Aude Oliva and Hyun Soo Park and James M. Rehg and Yoichi Sato and Jianbo Shi and Mike Zheng Shou and Antonio Torralba and Lorenzo Torresani and Mingfei Yan and Jitendra Malik},
      year={2022},
      eprint={2110.07058},
      archivePrefix={arXiv},
      primaryClass={cs.CV},
      url={https://arxiv.org/abs/2110.07058}, 
}

@misc{oncescu2021querydvideodatasethighquality,
      title={QuerYD: A video dataset with high-quality text and audio narrations}, 
      author={Andreea-Maria Oncescu and João F. Henriques and Yang Liu and Andrew Zisserman and Samuel Albanie},
      year={2021},
      eprint={2011.11071},
      archivePrefix={arXiv},
      primaryClass={cs.CV},
      url={https://arxiv.org/abs/2011.11071}, 
}

@misc{gupta2022datasetmedicalinstructionalvideo,
      title={A Dataset for Medical Instructional Video Classification and Question Answering}, 
      author={Deepak Gupta and Kush Attal and Dina Demner-Fushman},
      year={2022},
      eprint={2201.12888},
      archivePrefix={arXiv},
      primaryClass={cs.CV},
      url={https://arxiv.org/abs/2201.12888}, 
}

@misc{hendricks2017localizingmomentsvideonatural,
      title={Localizing Moments in Video with Natural Language}, 
      author={Lisa Anne Hendricks and Oliver Wang and Eli Shechtman and Josef Sivic and Trevor Darrell and Bryan Russell},
      year={2017},
      eprint={1708.01641},
      archivePrefix={arXiv},
      primaryClass={cs.CV},
      url={https://arxiv.org/abs/1708.01641}, 
}

@misc{zhang2025llavavideovideoinstructiontuning,
      title={LLaVA-Video: Video Instruction Tuning With Synthetic Data}, 
      author={Yuanhan Zhang and Jinming Wu and Wei Li and Bo Li and Zejun Ma and Ziwei Liu and Chunyuan Li},
      year={2025},
      eprint={2410.02713},
      archivePrefix={arXiv},
      primaryClass={cs.CV},
      url={https://arxiv.org/abs/2410.02713}, 
}

@misc{lei2019tvqalocalizedcompositionalvideo,
      title={TVQA: Localized, Compositional Video Question Answering}, 
      author={Jie Lei and Licheng Yu and Mohit Bansal and Tamara L. Berg},
      year={2019},
      eprint={1809.01696},
      archivePrefix={arXiv},
      primaryClass={cs.CL},
      url={https://arxiv.org/abs/1809.01696}, 
}

@misc{xiao2021nextqanextphasequestionansweringexplaining,
      title={NExT-QA:Next Phase of Question-Answering to Explaining Temporal Actions}, 
      author={Junbin Xiao and Xindi Shang and Angela Yao and Tat-Seng Chua},
      year={2021},
      eprint={2105.08276},
      archivePrefix={arXiv},
      primaryClass={cs.CV},
      url={https://arxiv.org/abs/2105.08276}, 
}

@misc{yi2020clevrercollisioneventsvideo,
      title={CLEVRER: CoLlision Events for Video REpresentation and Reasoning}, 
      author={Kexin Yi and Chuang Gan and Yunzhu Li and Pushmeet Kohli and Jiajun Wu and Antonio Torralba and Joshua B. Tenenbaum},
      year={2020},
      eprint={1910.01442},
      archivePrefix={arXiv},
      primaryClass={cs.CV},
      url={https://arxiv.org/abs/1910.01442}, 
}

@inproceedings{shao2020tapos,
title={Intra- and Inter-Action Understanding via Temporal Action Parsing},
author={Shao, Dian and Zhao, Yue and Dai, Bo and Lin, Dahua},
booktitle={IEEE Conference on Computer Vision and Pattern Recognition (CVPR)},
year={2020}
}

@misc{tang2021humancentricspatiotemporalvideogrounding,
      title={Human-centric Spatio-Temporal Video Grounding With Visual Transformers}, 
      author={Zongheng Tang and Yue Liao and Si Liu and Guanbin Li and Xiaojie Jin and Hongxu Jiang and Qian Yu and Dong Xu},
      year={2021},
      eprint={2011.05049},
      archivePrefix={arXiv},
      primaryClass={cs.CV},
      url={https://arxiv.org/abs/2011.05049}, 
}

@misc{bansal2022viewbestviewprocedure,
      title={My View is the Best View: Procedure Learning from Egocentric Videos}, 
      author={Siddhant Bansal and Chetan Arora and C. V. Jawahar},
      year={2022},
      eprint={2207.10883},
      archivePrefix={arXiv},
      primaryClass={cs.CV},
      url={https://arxiv.org/abs/2207.10883}, 
}

@misc{zhukov2019crosstaskweaklysupervisedlearning,
      title={Cross-task weakly supervised learning from instructional videos}, 
      author={Dimitri Zhukov and Jean-Baptiste Alayrac and Ramazan Gokberk Cinbis and David Fouhey and Ivan Laptev and Josef Sivic},
      year={2019},
      eprint={1903.08225},
      archivePrefix={arXiv},
      primaryClass={cs.CV},
      url={https://arxiv.org/abs/1903.08225}, 
}

@misc{wang2024mementoscomprehensivebenchmarkmultimodal,
      title={Mementos: A Comprehensive Benchmark for Multimodal Large Language Model Reasoning over Image Sequences}, 
      author={Xiyao Wang and Yuhang Zhou and Xiaoyu Liu and Hongjin Lu and Yuancheng Xu and Feihong He and Jaehong Yoon and Taixi Lu and Gedas Bertasius and Mohit Bansal and Huaxiu Yao and Furong Huang},
      year={2024},
      eprint={2401.10529},
      archivePrefix={arXiv},
      primaryClass={cs.CV},
      url={https://arxiv.org/abs/2401.10529}, 
}

@misc{karmanov2025eclairextractingcontent,
      title={\'Eclair -- Extracting Content and Layout with Integrated Reading Order for Documents}, 
      author={Ilia Karmanov and Amala Sanjay Deshmukh and Lukas Voegtle and Philipp Fischer and Kateryna Chumachenko and Timo Roman and Jarno Seppänen and Jupinder Parmar and Joseph Jennings and Andrew Tao and Karan Sapra},
      year={2025},
      eprint={2502.04223},
      archivePrefix={arXiv},
      primaryClass={cs.CV},
      url={https://arxiv.org/abs/2502.04223}, 
}

@inproceedings{PontTuset_eccv2020,
  author    = {Jordi Pont-Tuset and Jasper Uijlings and Soravit Changpinyo and Radu Soricut and Vittorio Ferrari},
  title     = {Connecting Vision and Language with Localized Narratives},
  booktitle = {ECCV},
  year      = {2020}
}

@article{Kuznetsova_2020,
   title={The Open Images Dataset V4: Unified Image Classification, Object Detection, and Visual Relationship Detection at Scale},
   volume={128},
   ISSN={1573-1405},
   url={http://dx.doi.org/10.1007/s11263-020-01316-z},
   DOI={10.1007/s11263-020-01316-z},
   number={7},
   journal={International Journal of Computer Vision},
   publisher={Springer Science and Business Media LLC},
   author={Kuznetsova, Alina and Rom, Hassan and Alldrin, Neil and Uijlings, Jasper and Krasin, Ivan and Pont-Tuset, Jordi and Kamali, Shahab and Popov, Stefan and Malloci, Matteo and Kolesnikov, Alexander and Duerig, Tom and Ferrari, Vittorio},
   year={2020},
   month=mar, pages={1956–1981} }

@misc{sidorov2020textcapsdatasetimagecaptioning,
      title={TextCaps: a Dataset for Image Captioning with Reading Comprehension}, 
      author={Oleksii Sidorov and Ronghang Hu and Marcus Rohrbach and Amanpreet Singh},
      year={2020},
      eprint={2003.12462},
      archivePrefix={arXiv},
      primaryClass={cs.CV},
      url={https://arxiv.org/abs/2003.12462}, 
}

@inproceedings{singh2019towards,
    title={Towards VQA Models That Can Read},
    author={Singh, Amanpreet and Natarjan, Vivek and Shah, Meet and Jiang, Yu and Chen, Xinlei and Parikh, Devi and Rohrbach, Marcus},
    booktitle={Proceedings of the IEEE Conference on Computer Vision and Pattern Recognition},
    pages={8317-8326},
    year={2019}
}

@misc{marino2019okvqavisualquestionanswering,
      title={OK-VQA: A Visual Question Answering Benchmark Requiring External Knowledge}, 
      author={Kenneth Marino and Mohammad Rastegari and Ali Farhadi and Roozbeh Mottaghi},
      year={2019},
      eprint={1906.00067},
      archivePrefix={arXiv},
      primaryClass={cs.CV},
      url={https://arxiv.org/abs/1906.00067}, 
}

@misc{hudson2019gqanewdatasetrealworld,
      title={GQA: A New Dataset for Real-World Visual Reasoning and Compositional Question Answering}, 
      author={Drew A. Hudson and Christopher D. Manning},
      year={2019},
      eprint={1902.09506},
      archivePrefix={arXiv},
      primaryClass={cs.CL},
      url={https://arxiv.org/abs/1902.09506}, 
}

@misc{johnson2016clevrdiagnosticdatasetcompositional,
      title={CLEVR: A Diagnostic Dataset for Compositional Language and Elementary Visual Reasoning}, 
      author={Justin Johnson and Bharath Hariharan and Laurens van der Maaten and Li Fei-Fei and C. Lawrence Zitnick and Ross Girshick},
      year={2016},
      eprint={1612.06890},
      archivePrefix={arXiv},
      primaryClass={cs.CV},
      url={https://arxiv.org/abs/1612.06890}, 
}

@misc{lindström2022clevrmathdatasetcompositionallanguage,
      title={CLEVR-Math: A Dataset for Compositional Language, Visual and Mathematical Reasoning}, 
      author={Adam Dahlgren Lindström and Savitha Sam Abraham},
      year={2022},
      eprint={2208.05358},
      archivePrefix={arXiv},
      primaryClass={cs.LG},
      url={https://arxiv.org/abs/2208.05358}, 
}

@misc{acharya2018tallyqaansweringcomplexcounting,
      title={{TallyQA}: Answering Complex Counting Questions}, 
      author={Manoj Acharya and Kushal Kafle and Christopher Kanan},
      year={2018},
      eprint={1810.12440},
      archivePrefix={arXiv},
      primaryClass={cs.CV},
      url={https://arxiv.org/abs/1810.12440}, 
}

@misc{DatabricksBlog2023DollyV2,
  author    = {Mike Conover and Matt Hayes and Ankit Mathur and Jianwei Xie and Jun Wan and Sam Shah and Ali Ghodsi and Patrick Wendell and Matei Zaharia and Reynold Xin},
  title     = {Free Dolly: Introducing the World's First Truly Open Instruction-Tuned LLM},
  year      = {2023},
  howpublished = {\url{https://www.databricks.com/blog/2023/04/12/dolly-first-open-commercially-viable-instruction-tuned-llm}}
}

@misc{hsiao2025screenqalargescalequestionanswerpairs,
      title={ScreenQA: Large-Scale Question-Answer Pairs over Mobile App Screenshots}, 
      author={Yu-Chung Hsiao and Fedir Zubach and Gilles Baechler and Srinivas Sunkara and Victor Carbune and Jason Lin and Maria Wang and Yun Zhu and Jindong Chen},
      year={2025},
      eprint={2209.08199},
      archivePrefix={arXiv},
      primaryClass={cs.CL},
      url={https://arxiv.org/abs/2209.08199}, 
}

@misc{gurari2018vizwizgrandchallengeanswering,
      title={VizWiz Grand Challenge: Answering Visual Questions from Blind People}, 
      author={Danna Gurari and Qing Li and Abigale J. Stangl and Anhong Guo and Chi Lin and Kristen Grauman and Jiebo Luo and Jeffrey P. Bigham},
      year={2018},
      eprint={1802.08218},
      archivePrefix={arXiv},
      primaryClass={cs.CV},
      url={https://arxiv.org/abs/1802.08218}, 
}

@inproceedings{kazemzadeh-etal-2014-referitgame,
    title = "{R}efer{I}t{G}ame: Referring to Objects in Photographs of Natural Scenes",
    author = "Kazemzadeh, Sahar  and
      Ordonez, Vicente  and
      Matten, Mark  and
      Berg, Tamara",
    editor = "Moschitti, Alessandro  and
      Pang, Bo  and
      Daelemans, Walter",
    booktitle = "Proceedings of the 2014 Conference on Empirical Methods in Natural Language Processing ({EMNLP})",
    month = oct,
    year = "2014",
    address = "Doha, Qatar",
    publisher = "Association for Computational Linguistics",
    url = "https://aclanthology.org/D14-1086",
    doi = "10.3115/v1/D14-1086",
    pages = "787--798",
}

@inproceedings{kim2022donut,
  title     = {OCR-Free Document Understanding Transformer},
  author    = {Kim, Geewook and Hong, Teakgyu and Yim, Moonbin and Nam, JeongYeon and Park, Jinyoung and Yim, Jinyeong and Hwang, Wonseok and Yun, Sangdoo and Han, Dongyoon and Park, Seunghyun},
  booktitle = {European Conference on Computer Vision (ECCV)},
  year      = {2022}
}

@misc{nassar2022tableformertablestructureunderstanding,
      title={TableFormer: Table Structure Understanding with Transformers}, 
      author={Ahmed Nassar and Nikolaos Livathinos and Maksym Lysak and Peter Staar},
      year={2022},
      eprint={2203.01017},
      archivePrefix={arXiv},
      primaryClass={cs.CV},
      url={https://arxiv.org/abs/2203.01017}, 
}

@misc{wang2020vatexlargescalehighqualitymultilingual,
      title={VATEX: A Large-Scale, High-Quality Multilingual Dataset for Video-and-Language Research}, 
      author={Xin Wang and Jiawei Wu and Junkun Chen and Lei Li and Yuan-Fang Wang and William Yang Wang},
      year={2020},
      eprint={1904.03493},
      archivePrefix={arXiv},
      primaryClass={cs.CV},
      url={https://arxiv.org/abs/1904.03493}, 
}

@misc{deitke2024molmopixmoopenweights,
      title={Molmo and PixMo: Open Weights and Open Data for State-of-the-Art Vision-Language Models}, 
      author={Matt Deitke and Christopher Clark and Sangho Lee and Rohun Tripathi and Yue Yang and Jae Sung Park and Mohammadreza Salehi and Niklas Muennighoff and Kyle Lo and Luca Soldaini and Jiasen Lu and Taira Anderson and Erin Bransom and Kiana Ehsani and Huong Ngo and YenSung Chen and Ajay Patel and Mark Yatskar and Chris Callison-Burch and Andrew Head and Rose Hendrix and Favyen Bastani and Eli VanderBilt and Nathan Lambert and Yvonne Chou and Arnavi Chheda and Jenna Sparks and Sam Skjonsberg and Michael Schmitz and Aaron Sarnat and Byron Bischoff and Pete Walsh and Chris Newell and Piper Wolters and Tanmay Gupta and Kuo-Hao Zeng and Jon Borchardt and Dirk Groeneveld and Crystal Nam and Sophie Lebrecht and Caitlin Wittlif and Carissa Schoenick and Oscar Michel and Ranjay Krishna and Luca Weihs and Noah A. Smith and Hannaneh Hajishirzi and Ross Girshick and Ali Farhadi and Aniruddha Kembhavi},
      year={2024},
      eprint={2409.17146},
      archivePrefix={arXiv},
      primaryClass={cs.CV},
      url={https://arxiv.org/abs/2409.17146}, 
}

@misc{goyal2017makingvvqamatter,
      title={Making the V in VQA Matter: Elevating the Role of Image Understanding in Visual Question Answering}, 
      author={Yash Goyal and Tejas Khot and Douglas Summers-Stay and Dhruv Batra and Devi Parikh},
      year={2017},
      eprint={1612.00837},
      archivePrefix={arXiv},
      primaryClass={cs.CV},
      url={https://arxiv.org/abs/1612.00837}, 
}

@inproceedings{shridhar2020alfred,
  title={Alfred: A benchmark for interpreting grounded instructions for everyday tasks},
  author={Shridhar, Mohit and Thomason, Jesse and Gordon, Daniel and Bisk, Yonatan and Han, Winson and Mottaghi, Roozbeh and Zettlemoyer, Luke and Fox, Dieter},
  booktitle={Proceedings of the IEEE/CVF conference on computer vision and pattern recognition},
  pages={10740--10749},
  year={2020}
}

@article{Maaz2024VideoGPT+,
      title={VideoGPT+: Integrating Image and Video Encoders for Enhanced Video Understanding},
      author={Maaz, Muhammad and Rasheed, Hanoona and Khan, Salman and Khan, Fahad Shahbaz},
      journal={arxiv},
      year={2024},
      url={https://arxiv.org/abs/2406.09418}
  }

@misc{long2022endtoendunifiedscenetext,
      title={Towards End-to-End Unified Scene Text Detection and Layout Analysis}, 
      author={Shangbang Long and Siyang Qin and Dmitry Panteleev and Alessandro Bissacco and Yasuhisa Fujii and Michalis Raptis},
      year={2022},
      eprint={2203.15143},
      archivePrefix={arXiv},
      primaryClass={cs.CV},
      url={https://arxiv.org/abs/2203.15143}, 
}

@misc{chen2024rightwayevaluatinglarge,
      title={Are We on the Right Way for Evaluating Large Vision-Language Models?}, 
      author={Lin Chen and Jinsong Li and Xiaoyi Dong and Pan Zhang and Yuhang Zang and Zehui Chen and Haodong Duan and Jiaqi Wang and Yu Qiao and Dahua Lin and Feng Zhao},
      year={2024},
      eprint={2403.20330},
      archivePrefix={arXiv},
      primaryClass={cs.CV},
      url={https://arxiv.org/abs/2403.20330}, 
}

@article{fu2024blink,
          title={BLINK: Multimodal Large Language Models Can See but Not Perceive},
          author={Fu, Xingyu and Hu, Yushi and Li, Bangzheng and Feng, Yu and Wang, Haoyu and Lin, Xudong and Roth, Dan and Smith, Noah A and Ma, Wei-Chiu and Krishna, Ranjay},
          journal={arXiv preprint arXiv:2404.12390},
          year={2024}
        }

@misc{wang2024muirbenchcomprehensivebenchmarkrobust,
      title={MuirBench: A Comprehensive Benchmark for Robust Multi-image Understanding}, 
      author={Fei Wang and Xingyu Fu and James Y. Huang and Zekun Li and Qin Liu and Xiaogeng Liu and Mingyu Derek Ma and Nan Xu and Wenxuan Zhou and Kai Zhang and Tianyi Lorena Yan and Wenjie Jacky Mo and Hsiang-Hui Liu and Pan Lu and Chunyuan Li and Chaowei Xiao and Kai-Wei Chang and Dan Roth and Sheng Zhang and Hoifung Poon and Muhao Chen},
      year={2024},
      eprint={2406.09411},
      archivePrefix={arXiv},
      primaryClass={cs.CV},
      url={https://arxiv.org/abs/2406.09411}, 
}

@misc{guan2024hallusionbenchadvanceddiagnosticsuite,
      title={HallusionBench: An Advanced Diagnostic Suite for Entangled Language Hallucination and Visual Illusion in Large Vision-Language Models}, 
      author={Tianrui Guan and Fuxiao Liu and Xiyang Wu and Ruiqi Xian and Zongxia Li and Xiaoyu Liu and Xijun Wang and Lichang Chen and Furong Huang and Yaser Yacoob and Dinesh Manocha and Tianyi Zhou},
      year={2024},
      eprint={2310.14566},
      archivePrefix={arXiv},
      primaryClass={cs.CV},
      url={https://arxiv.org/abs/2310.14566}, 
}

@article{roberts2025zerobench,
  title={Zerobench: An impossible visual benchmark for contemporary large multimodal models},
  author={Roberts, Jonathan and Taesiri, Mohammad Reza and Sharma, Ansh and Gupta, Akash and Roberts, Samuel and Croitoru, Ioana and Bogolin, Simion-Vlad and Tang, Jialu and Langer, Florian and Raina, Vyas and others},
  journal={arXiv preprint arXiv:2502.09696},
  year={2025}
}

@misc{wang2024allseeingprojectv2general,
      title={The All-Seeing Project V2: Towards General Relation Comprehension of the Open World}, 
      author={Weiyun Wang and Yiming Ren and Haowen Luo and Tiantong Li and Chenxiang Yan and Zhe Chen and Wenhai Wang and Qingyun Li and Lewei Lu and Xizhou Zhu and Yu Qiao and Jifeng Dai},
      year={2024},
      eprint={2402.19474},
      archivePrefix={arXiv},
      primaryClass={cs.CV},
      url={https://arxiv.org/abs/2402.19474}, 
}

@misc{li2023evaluatingobjecthallucinationlarge,
      title={Evaluating Object Hallucination in Large Vision-Language Models}, 
      author={Yifan Li and Yifan Du and Kun Zhou and Jinpeng Wang and Wayne Xin Zhao and Ji-Rong Wen},
      year={2023},
      eprint={2305.10355},
      archivePrefix={arXiv},
      primaryClass={cs.CV},
      url={https://arxiv.org/abs/2305.10355}, 
}

@misc{zhang2025mmerealworldmultimodalllmchallenge,
      title={MME-RealWorld: Could Your Multimodal LLM Challenge High-Resolution Real-World Scenarios that are Difficult for Humans?}, 
      author={Yi-Fan Zhang and Huanyu Zhang and Haochen Tian and Chaoyou Fu and Shuangqing Zhang and Junfei Wu and Feng Li and Kun Wang and Qingsong Wen and Zhang Zhang and Liang Wang and Rong Jin and Tieniu Tan},
      year={2025},
      eprint={2408.13257},
      archivePrefix={arXiv},
      primaryClass={cs.CV},
      url={https://arxiv.org/abs/2408.13257}, 
}

@misc{ying2024mmtbenchcomprehensivemultimodalbenchmark,
      title={MMT-Bench: A Comprehensive Multimodal Benchmark for Evaluating Large Vision-Language Models Towards Multitask AGI}, 
      author={Kaining Ying and Fanqing Meng and Jin Wang and Zhiqian Li and Han Lin and Yue Yang and Hao Zhang and Wenbo Zhang and Yuqi Lin and Shuo Liu and Jiayi Lei and Quanfeng Lu and Runjian Chen and Peng Xu and Renrui Zhang and Haozhe Zhang and Peng Gao and Yali Wang and Yu Qiao and Ping Luo and Kaipeng Zhang and Wenqi Shao},
      year={2024},
      eprint={2404.16006},
      archivePrefix={arXiv},
      primaryClass={cs.CV},
      url={https://arxiv.org/abs/2404.16006}, 
}

@misc{guo2025rbenchgraduatelevelmultidisciplinarybenchmarks,
      title={R-Bench: Graduate-level Multi-disciplinary Benchmarks for LLM \& MLLM Complex Reasoning Evaluation}, 
      author={Meng-Hao Guo and Jiajun Xu and Yi Zhang and Jiaxi Song and Haoyang Peng and Yi-Xuan Deng and Xinzhi Dong and Kiyohiro Nakayama and Zhengyang Geng and Chen Wang and Bolin Ni and Guo-Wei Yang and Yongming Rao and Houwen Peng and Han Hu and Gordon Wetzstein and Shi-min Hu},
      year={2025},
      eprint={2505.02018},
      archivePrefix={arXiv},
      primaryClass={cs.CV},
      url={https://arxiv.org/abs/2505.02018}, 
}

@article{lu2024wildvision,
  title={Wildvision: Evaluating vision-language models in the wild with human preferences},
  author={Lu, Yujie and Jiang, Dongfu and Chen, Wenhu and Wang, William Yang and Choi, Yejin and Lin, Bill Yuchen},
  journal={Advances in Neural Information Processing Systems},
  volume={37},
  pages={48224--48255},
  year={2024}
}

@inproceedings{yue2023mmmu,
            title={MMMU: A Massive Multi-discipline Multimodal Understanding and Reasoning Benchmark for Expert AGI},
            author={Xiang Yue and Yuansheng Ni and Kai Zhang and Tianyu Zheng and Ruoqi Liu and Ge Zhang and Samuel Stevens and Dongfu Jiang and Weiming Ren and Yuxuan Sun and Cong Wei and Botao Yu and Ruibin Yuan and Renliang Sun and Ming Yin and Boyuan Zheng and Zhenzhu Yang and Yibo Liu and Wenhao Huang and Huan Sun and Yu Su and Wenhu Chen},
            booktitle={Proceedings of CVPR},
            year={2024},
          }

@misc{yue2025mmmuprorobustmultidisciplinemultimodal,
      title={MMMU-Pro: A More Robust Multi-discipline Multimodal Understanding Benchmark}, 
      author={Xiang Yue and Tianyu Zheng and Yuansheng Ni and Yubo Wang and Kai Zhang and Shengbang Tong and Yuxuan Sun and Botao Yu and Ge Zhang and Huan Sun and Yu Su and Wenhu Chen and Graham Neubig},
      year={2025},
      eprint={2409.02813},
      archivePrefix={arXiv},
      primaryClass={cs.CL},
      url={https://arxiv.org/abs/2409.02813}, 
}

@inproceedings{lu2024mathvista,
  author    = {Lu, Pan and Bansal, Hritik and Xia, Tony and Liu, Jiacheng and Li, Chunyuan and Hajishirzi, Hannaneh and Cheng, Hao and Chang, Kai-Wei and Galley, Michel and Gao, Jianfeng},
  title     = {MathVista: Evaluating Mathematical Reasoning of Foundation Models in Visual Contexts},
  booktitle={International Conference on Learning Representations (ICLR)},
  year      = {2024}
}

@inproceedings{
wang2024measuring,
title={Measuring Multimodal Mathematical Reasoning with MATH-Vision Dataset},
author={Ke Wang and Junting Pan and Weikang Shi and Zimu Lu and Houxing Ren and Aojun Zhou and Mingjie Zhan and Hongsheng Li},
booktitle={The Thirty-eight Conference on Neural Information Processing Systems Datasets and Benchmarks Track},
year={2024},
url={https://openreview.net/forum?id=QWTCcxMpPA}
}

@misc{masry2022chartqabenchmarkquestionanswering,
      title={ChartQA: A Benchmark for Question Answering about Charts with Visual and Logical Reasoning}, 
      author={Ahmed Masry and Do Xuan Long and Jia Qing Tan and Shafiq Joty and Enamul Hoque},
      year={2022},
      eprint={2203.10244},
      archivePrefix={arXiv},
      primaryClass={cs.CL},
      url={https://arxiv.org/abs/2203.10244}, 
}

@misc{mathew2021infographicvqa,
      title={InfographicVQA}, 
      author={Minesh Mathew and Viraj Bagal and Rubèn Pérez Tito and Dimosthenis Karatzas and Ernest Valveny and C. V Jawahar},
      year={2021},
      eprint={2104.12756},
      archivePrefix={arXiv},
      primaryClass={cs.CV},
      url={https://arxiv.org/abs/2104.12756}, 
}

@misc{kembhavi2016diagramworthdozenimages,
      title={A Diagram Is Worth A Dozen Images}, 
      author={Aniruddha Kembhavi and Mike Salvato and Eric Kolve and Minjoon Seo and Hannaneh Hajishirzi and Ali Farhadi},
      year={2016},
      eprint={1603.07396},
      archivePrefix={arXiv},
      primaryClass={cs.CV},
      url={https://arxiv.org/abs/1603.07396}, 
}

@misc{mathew2021docvqadatasetvqadocument,
      title={DocVQA: A Dataset for VQA on Document Images}, 
      author={Minesh Mathew and Dimosthenis Karatzas and C. V. Jawahar},
      year={2021},
      eprint={2007.00398},
      archivePrefix={arXiv},
      primaryClass={cs.CV},
      url={https://arxiv.org/abs/2007.00398}, 
}

@misc{zhu2016visual7wgroundedquestionanswering,
      title={Visual7W: Grounded Question Answering in Images}, 
      author={Yuke Zhu and Oliver Groth and Michael Bernstein and Li Fei-Fei},
      year={2016},
      eprint={1511.03416},
      archivePrefix={arXiv},
      primaryClass={cs.CV},
      url={https://arxiv.org/abs/1511.03416}, 
}

@inproceedings{Pfitzmann_2022, series={KDD ’22},
   title={DocLayNet: A Large Human-Annotated Dataset for Document-Layout Segmentation},
   url={http://dx.doi.org/10.1145/3534678.3539043},
   DOI={10.1145/3534678.3539043},
   booktitle={Proceedings of the 28th ACM SIGKDD Conference on Knowledge Discovery and Data Mining},
   publisher={ACM},
   author={Pfitzmann, Birgit and Auer, Christoph and Dolfi, Michele and Nassar, Ahmed S. and Staar, Peter},
   year={2022},
   month=aug, pages={3743–3751},
   collection={KDD ’22} }

@article{tang2020multilingual,
  title={Multilingual translation with extensible multilingual pretraining and finetuning},
  author={Tang, Yuqing and Tran, Chau and Li, Xian and Chen, Peng-Jen and Goyal, Naman and Chaudhary, Vishrav and Gu, Jiatao and Fan, Angela},
  journal={arXiv preprint arXiv:2008.00401},
  year={2020}
}

@misc{laurençon2024unlockingconversionwebscreenshots,
      title={Unlocking the conversion of Web Screenshots into HTML Code with the WebSight Dataset}, 
      author={Hugo Laurençon and Léo Tronchon and Victor Sanh},
      year={2024},
      eprint={2403.09029},
      archivePrefix={arXiv},
      primaryClass={cs.HC},
      url={https://arxiv.org/abs/2403.09029}, 
}

@article{yang2023large,
  title={A large-scale dataset for end-to-end table recognition in the wild},
  author={Yang, Fan and Hu, Lei and Liu, Xinwu and Huang, Shuangping and Gu, Zhenghui},
  journal={Scientific Data},
  volume={10},
  number={1},
  pages={110},
  year={2023},
  publisher={Nature Publishing Group UK London}
}

@misc{zheng2020globaltableextractorgte,
      title={Global Table Extractor (GTE): A Framework for Joint Table Identification and Cell Structure Recognition Using Visual Context}, 
      author={Xinyi Zheng and Doug Burdick and Lucian Popa and Xu Zhong and Nancy Xin Ru Wang},
      year={2020},
      eprint={2005.00589},
      archivePrefix={arXiv},
      primaryClass={cs.CV},
      url={https://arxiv.org/abs/2005.00589}, 
}

@misc{smock2021pubtables1mcomprehensivetableextraction,
      title={PubTables-1M: Towards comprehensive table extraction from unstructured documents}, 
      author={Brandon Smock and Rohith Pesala and Robin Abraham},
      year={2021},
      eprint={2110.00061},
      archivePrefix={arXiv},
      primaryClass={cs.LG},
      url={https://arxiv.org/abs/2110.00061}, 
}

@inproceedings{singh2021textocr,
  title={Textocr: Towards large-scale end-to-end reasoning for arbitrary-shaped scene text},
  author={Singh, Amanpreet and Pang, Guan and Toh, Mandy and Huang, Jing and Galuba, Wojciech and Hassner, Tal},
  booktitle={Proceedings of the IEEE/CVF conference on computer vision and pattern recognition},
  pages={8802--8812},
  year={2021}
}

@misc{jaume2019funsddatasetformunderstanding,
      title={FUNSD: A Dataset for Form Understanding in Noisy Scanned Documents}, 
      author={Guillaume Jaume and Hazim Kemal Ekenel and Jean-Philippe Thiran},
      year={2019},
      eprint={1905.13538},
      archivePrefix={arXiv},
      primaryClass={cs.IR},
      url={https://arxiv.org/abs/1905.13538}, 
}

@inproceedings{liu2011casia,
  title={CASIA online and offline Chinese handwriting databases},
  author={Liu, Cheng-Lin and Yin, Fei and Wang, Da-Han and Wang, Qiu-Feng},
  booktitle={2011 international conference on document analysis and recognition},
  pages={37--41},
  year={2011},
  organization={IEEE}
}

@misc{shi2018icdar2017competitionreadingchinese,
      title={ICDAR2017 Competition on Reading Chinese Text in the Wild (RCTW-17)}, 
      author={Baoguang Shi and Cong Yao and Minghui Liao and Mingkun Yang and Pei Xu and Linyan Cui and Serge Belongie and Shijian Lu and Xiang Bai},
      year={2018},
      eprint={1708.09585},
      archivePrefix={arXiv},
      primaryClass={cs.CV},
      url={https://arxiv.org/abs/1708.09585}, 
}

@misc{liu2019icdar2019robustreading,
      title={ICDAR 2019 Robust Reading Challenge on Reading Chinese Text on Signboard}, 
      author={Xi Liu and Rui Zhang and Yongsheng Zhou and Qianyi Jiang and Qi Song and Nan Li and Kai Zhou and Lei Wang and Dong Wang and Minghui Liao and Mingkun Yang and Xiang Bai and Baoguang Shi and Dimosthenis Karatzas and Shijian Lu and C. V. Jawahar},
      year={2019},
      eprint={1912.09641},
      archivePrefix={arXiv},
      primaryClass={cs.CV},
      url={https://arxiv.org/abs/1912.09641}, 
}

@misc{chang2022mapqadatasetquestionanswering,
      title={MapQA: A Dataset for Question Answering on Choropleth Maps}, 
      author={Shuaichen Chang and David Palzer and Jialin Li and Eric Fosler-Lussier and Ningchuan Xiao},
      year={2022},
      eprint={2211.08545},
      archivePrefix={arXiv},
      primaryClass={cs.CV},
      url={https://arxiv.org/abs/2211.08545}, 
}

@misc{agentsea_waveui25k_2024,
  title = {WaveUI-25k},
  author = {AgentSea},
  howpublished = {\url{https://huggingface.co/datasets/agentsea/wave-ui-25k}},
  year = {2024}
}
\bibliographystyle{references}

\end{document}